%% file: iclr2026_conference.tex
\documentclass{article} % For LaTeX2e
\usepackage{iclr2026_conference,times}

\input{math_commands.tex}

\usepackage[raster,skins, most]{tcolorbox}

\usepackage{hyperref}
\usepackage{url}
\usepackage{pifont}
\usepackage{booktabs}
\usepackage{xspace}
\usepackage{enumitem}
\usepackage{algpseudocodex}
\usepackage{algorithm}
\usepackage{multirow}
\usepackage[table,xcdraw]{xcolor}
\usepackage{graphicx}
\usepackage{subcaption} 
\usepackage{wrapfig}
\usepackage{listings}

\lstdefinestyle{lean}{
  language=,
  basicstyle=\ttfamily,
  keepspaces=true,
  columns=flexible,
  breaklines=true,
  showstringspaces=false,
  literate={∀}{$\forall$}{1}
           {ℝ}{$\mathbb{R}$}{1}
           {≤}{$\leq$}{1}
           {:=}{$:=$}{2}
           {→}{$\rightarrow$}{1}
           {λ}{$\lambda$}{1}
           {∧}{$\land$}{1}
}

\definecolor{darkgreen}{RGB}{50,100,0}
\definecolor{darkred}{RGB}{200, 0, 0}

\newcommand{\huggingface}{\raisebox{-1.5pt}{\includegraphics[height=1.05em]{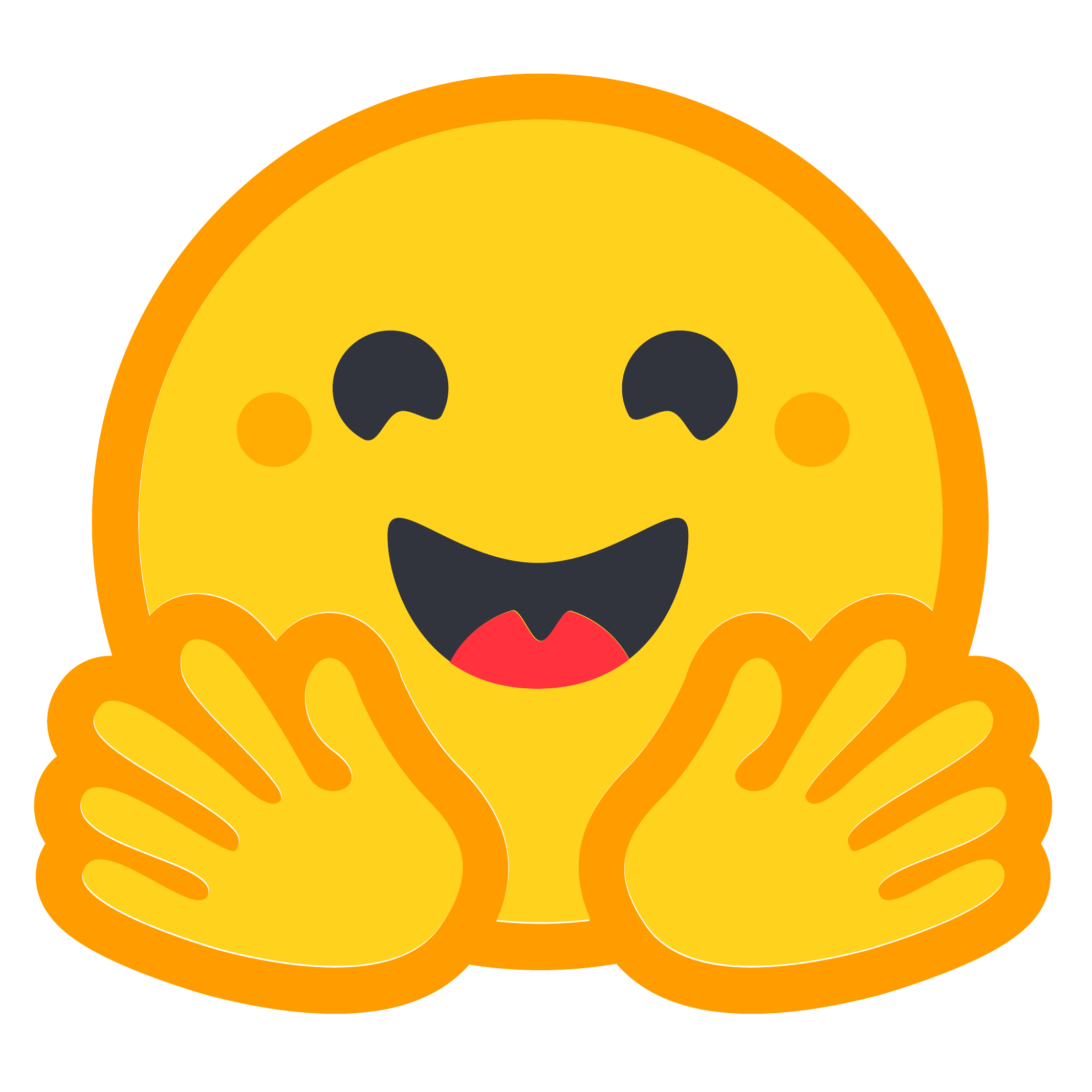}}\xspace}
\newcommand{\hfdataset}{\raisebox{-1.5pt}{\includegraphics[height=1.05em]{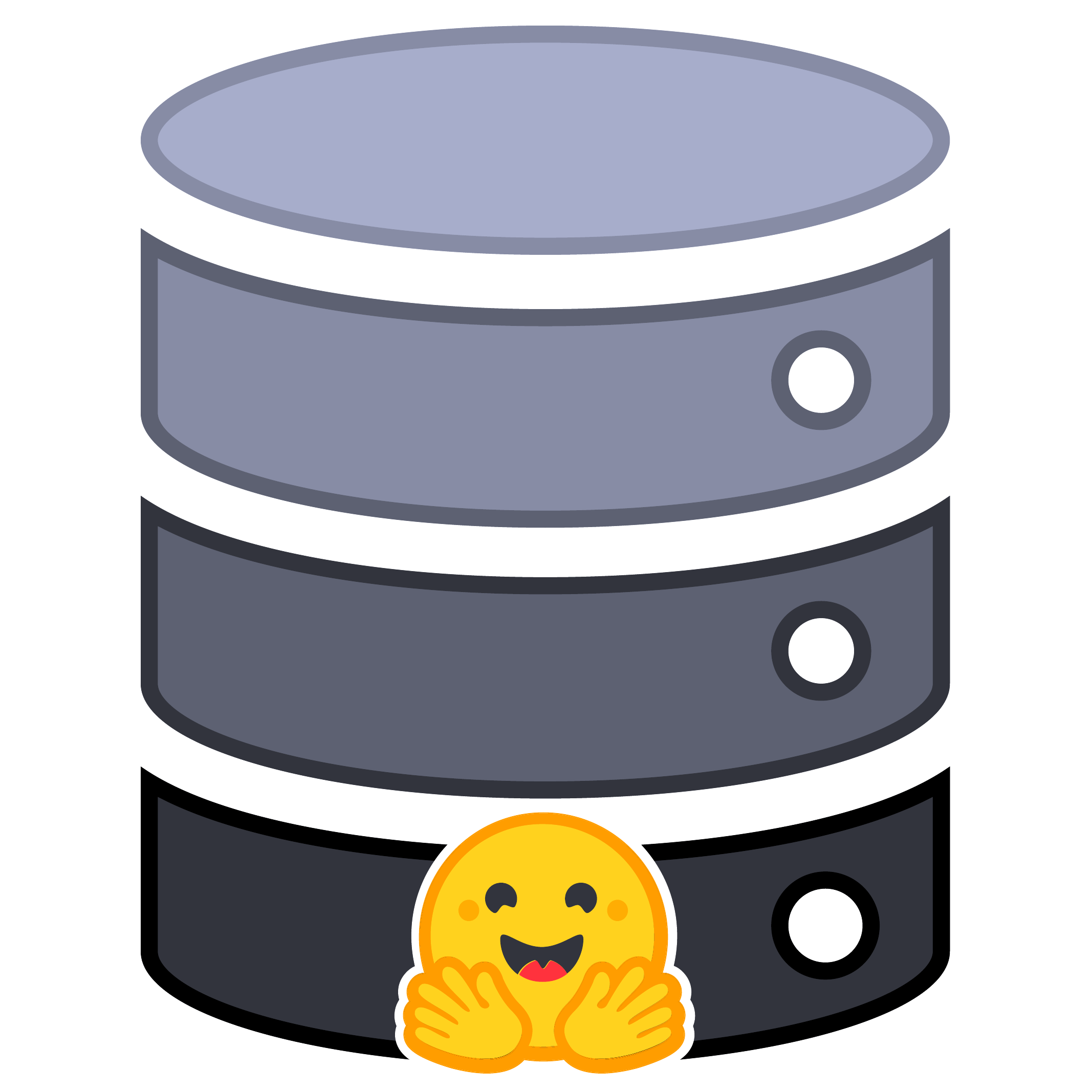}}\xspace}
\newcommand{\github}{\raisebox{-1.5pt}{\includegraphics[height=1.05em]{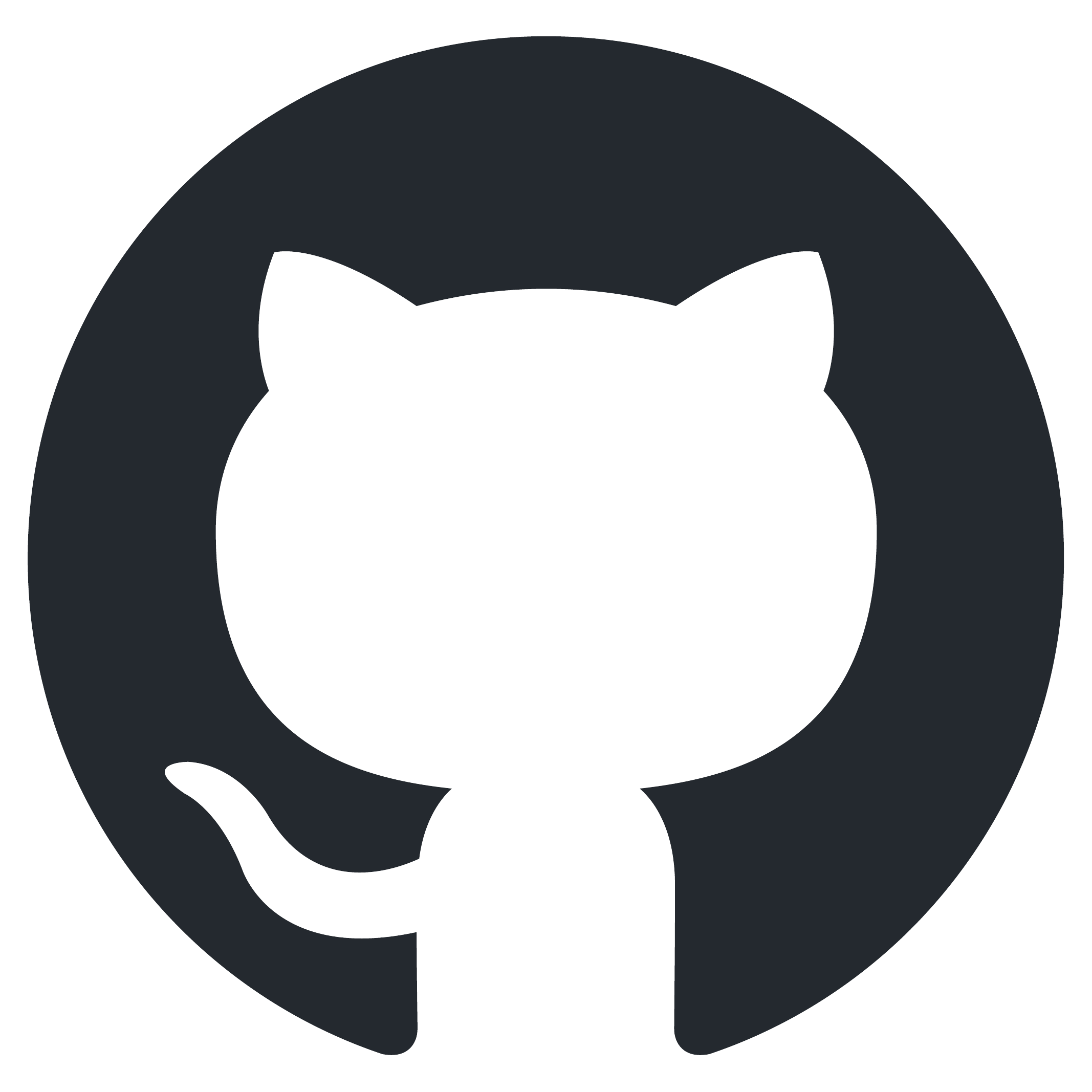}}\xspace}

\newcommand{\modelname}{\textsc{ReForm}\xspace}

\title{\modelname: Reflective Autoformalization with Prospective Bounded Sequence Optimization}

\iclrfinalcopy

\author{Guoxin Chen$^{1,4,\ast}$, Jing Wu$^{2,\ast,\dagger}$, Xinjie Chen$^{2,3}$, Wayne Xin Zhao$^{1,4,\dagger}$, Ruihua Song$^{1,\dagger}$\\
\textbf{Chengxi Li$^2$, Kai Fan$^2$, Dayiheng Liu$^2$, Minpeng Liao$^{2,\dagger}$}\\
$^1$Gaoling School of Artificial Intelligence, Renmin University of China\\
$^2$Tongyi Lab, Alibaba Group, $^3$Zhejiang University\\
$^4$Beijing Key Laboratory of Research on Large Models and Intelligent Governance\\
\texttt{\{gx.chen.chn, batmanfly\}@gmail.com, songruihua\_bloon@outlook.com}\\
\texttt{\{lz.wujing, chenxinjie.cxj, minpeng.lmp\}@alibaba-inc.com}\\
\centerline{
\hspace{.5em}\huggingface \href{https://huggingface.co/collections/GuoxinChen/reform}{Model}
\hspace{.5em}\hfdataset \href{https://huggingface.co/datasets/GuoxinChen/ConsistencyCheck}{Benchmark} 
\hspace{.5em}\github \href{https://github.com/Chen-GX/ReForm}{Code}
}
}

\begin{document}

\maketitle
\begingroup
  \renewcommand\thefootnote{}  % 清空脚注编号
  \footnotetext{$^*$ Equal Contribution. $^\dagger$ Corresponding Authors.}
\endgroup

\vspace{-1.5em}
\begin{abstract}
\vspace{-0.5em}
Autoformalization, which translates natural language mathematics into machine-verifiable formal statements, is critical for using formal mathematical reasoning to solve math problems stated in natural language.
While Large Language Models can generate syntactically correct formal statements, they often fail to preserve the original problem's semantic intent.
This limitation arises from the LLM approaches' treating autoformalization as a simplistic translation task which lacks mechanisms for self-reflection and iterative refinement that human experts naturally employ.
To address these issues, we propose ReForm, a Reflective Autoformalization method that tightly integrates semantic consistency evaluation into the autoformalization process.
This enables the model to iteratively generate formal statements, assess its semantic fidelity, and self-correct identified errors through progressive refinement.
To effectively train this reflective model, we introduce Prospective Bounded Sequence Optimization (PBSO), which employs different rewards at different sequence positions to ensure that the model develops both accurate autoformalization and correct semantic validations, preventing superficial critiques that would undermine the purpose of reflection.
Extensive experiments across four autoformalization benchmarks demonstrate that ReForm achieves an average improvement of 22.6 percentage points over the strongest baselines.
To further ensure evaluation reliability, we introduce ConsistencyCheck, a benchmark of 859 expert-annotated items that not only validates LLMs as judges but also reveals that autoformalization is inherently difficult: even human experts produce semantic errors in up to 38.5\% of cases.
\vspace{-2em}
\end{abstract}

\begin{figure}[h]
    \centering
    \includegraphics[width=0.95\linewidth]{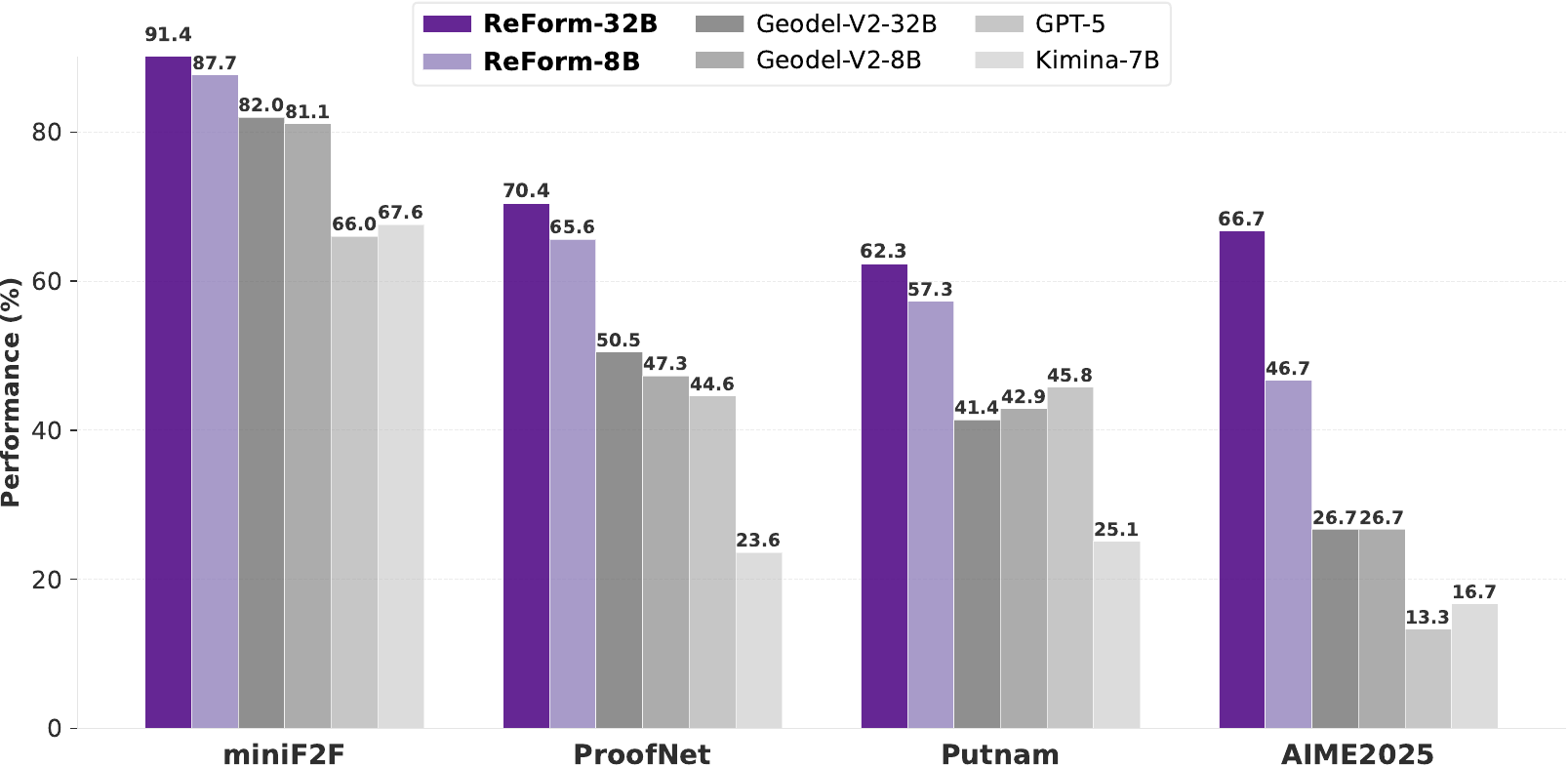}
    \vspace{-0.5em}
    \caption{Autoformalization performance of \modelname against state-of-the-art models.}
    % Autoformalization performance with both syntactic correctness and semantic consistency.
    \label{fig:abs_fig}
\end{figure}

\input{section/1.intro}
\input{section/2.related_work}
\input{section/3.method}
\input{section/4.exp}
\input{section/5.conclusion}

\clearpage

\subsubsection*{Acknowledgments}
This work was supported by Alibaba Research Intern Program.
This paper was partially supported by the National Natural Science Foundation of China No. 92470205 and Beijing Major Science and Technology Project under Contract No. Z251100008425002.

\subsubsection*{Ethics Statement}

Our work focuses on advancing automated mathematical formalization through a novel reflective paradigm and reinforcement learning algorithm. We have carefully considered the ethical implications of our research and taken appropriate measures to ensure responsible development.

\textbf{Human Annotation and Labor.} The construction of our ConsistencyCheck benchmark involved human experts in mathematics and Lean4. All annotators were fairly compensated for their expertise and time according to prevailing standards for skilled technical work. The annotation process was designed to be intellectually engaging rather than repetitive, leveraging the annotators' mathematical expertise. We ensured reasonable working conditions with no excessive time pressures, and annotators retained the right to decline or withdraw from tasks at any point.

\textbf{Intellectual Property and Attribution.} Our benchmark builds upon existing mathematical datasets (miniF2F and ProofNet), which we use in accordance with their licenses and with proper attribution. We acknowledge the substantial human effort that went into creating these original resources and ensure all sources are appropriately cited.

\textbf{Potential Impacts.} While our work aims to democratize access to formal mathematical reasoning tools, we acknowledge potential concerns. The automation of mathematical formalization could reduce demand for certain types of mathematical verification work. However, we believe our technology will primarily augment rather than replace human mathematicians, enabling them to focus on higher-level creative and conceptual work. Furthermore, by making formal verification more accessible, our work could enhance mathematical education and research, particularly in resource-constrained settings.

% \subsubsection*{Reproducibility Statement}

% To ensure the reproducibility of our work, we have made comprehensive efforts to document all aspects of our methodology and experiments. Section~\ref{subsec:setup} provides detailed descriptions of our evaluation benchmarks, baseline methods, and training data sources. Appendix~\ref{app:imp_details} contains thorough documentation of our data collection process for SFT trajectory data, along with complete hyperparameters for both SFT and RL training phases.
% Additionally, we provide extensive \textbf{supplementary materials} including our complete codebase with detailed implementation notes and example of our ConsistencyCheck benchmark, ensuring that researchers can readily reproduce our results.
% The entire ConsistencyCheck benchmark with expert annotations will be made publicly available upon acceptance.
% These materials collectively enable full reproduction of our experimental results and facilitate future research building upon our reflective autoformalization paradigm.

\bibliography{iclr2026_conference}
\bibliographystyle{iclr2026_conference}

\input{section/appendix}

% \appendix
% \section{Appendix}

\end{document}

%% file: math_commands.tex
\usepackage{amsmath,amsfonts,bm}

\def\eqref#1{equation~\ref{#1}}
\def\1{\bm{1}}

\DeclareMathAlphabet{\mathsfit}{\encodingdefault}{\sfdefault}{m}{sl}
\SetMathAlphabet{\mathsfit}{bold}{\encodingdefault}{\sfdefault}{bx}{n}

%% file: section/1.intro.tex
\section{Introduction}
\vspace{-1em}
Recent advances in Formal Mathematical Reasoning have demonstrated remarkable capabilities across a variety of challenging scenarios~\citep{polu2020generative,yang2023leandojo,xin2024deepseek,xin2024deepseekb,team2023internlm,wu2024internlm2,li2024survey,wu2025internlm2,ren2025deepseek,lin2025goedel,weng2025autoformalization}.
However, these advances are unevenly distributed across two symbiotic tasks: Automated Theorem Proving (ATP), the process of finding a proof for a given formal statement, and Autoformalization, the translation of natural language mathematical problems into formal, machine-verifiable statements such as those in Lean~\citep{de2015lean}.
This disparity establishes autoformalization as a critical bottleneck: it remains a labor-intensive endeavor that poses a great challenge even for human experts~\citep{chen2025seed}.

While Large Language Models (LLMs)~\citep{gpt_4o,claude_37,yang2025qwen3,gemini_25_pro} have shown proficiency in generating syntactically well-formed statements that pass Lean compiler verification (\textbf{syntactic correctness}), they often struggle to faithfully capture the semantic intent of the original problem (\textbf{semantic consistency}).
Recent studies~\citep{peng2025criticlean} have highlighted the pervasive issue of poor semantic fidelity in the current autoformalization systems.
Several concurrent works~\citep{wang2025kimina,lin2025goedelv2} have attempted to address this challenge by curating high-quality datasets specifically designed to improve semantic consistency while still treating autoformalization as a direct translation task in which models generate formal statements in a single forward pass, an approach we term the \textbf{\textit{one-pass generation paradigm}}.
While these data-centric efforts yield notable improvements, we find that models trained under this paradigm still frequently fail on subtle semantic details, such as misinterpreting quantifier scopes, overlooking implicit constraints, incorrectly formalizing edge cases, etc., that fundamentally compromise the original problem's intended meaning.
In this work, we argue that the root of this persistent limitation lies not only in the data quality, but more fundamentally in the one-pass generation paradigm itself: without any mechanism for self-reflection and correction, models cannot progressively identify and resolve their own semantic errors during generation.
This stands in stark contrast to how human experts tackle autoformalization. They employ an iterative process of review and refinement, continuously validating and adjusting their formal statements to ensure semantic fidelity.

Inspired by this, we propose \textbf{\modelname}, a novel \textbf{\textit{Reflective Autoformalization paradigm}} that emulates the human process of iterative review and refinement to enhance semantic consistency.
Instead of treating autoformalization as a single-pass translation task, \modelname reconceptualizes it as a reflective, iterative process that interweaves autoformalization with semantic self-validation.
Specifically, \modelname operates through a self-correction loop: (1) it first generates a candidate formal statement, (2) then critically evaluates whether this formalization faithfully captures the original problem's semantics, and (3) iteratively refines the statement based on the identified semantic discrepancies.
Unlike traditional one-pass approaches that commit to a single translation, this reflective paradigm enables the model to detect and correct its own semantic errors during the generation process, significantly reducing the risk of meaning distortions.

To prevent superficial or hallucinated critiques in the semantic self-validation, we design a \textbf{\textit{heterogeneous reward mechanism}} that targets two synergistic objectives: achieving correct final formal statements as the primary task ($r_{\text{task}}$ rewarded at the sequence end) and producing accurate semantic validation critiques as the auxiliary task ($r_{\text{aux}}$ rewarded at intermediate steps).
However, optimizing such heterogeneous rewards poses a significant challenge for existing reinforcement learning (RL) methods, which typically handle only a single terminal reward.
We therefore introduce \textbf{\textit{Prospective Bounded Sequence Optimization (PBSO)}}, a novel RL algorithm that enables optimizing multiple reward signals at different sequence positions.
The key innovation of PBSO lies in its \textit{prospective bounded return}, which smoothly integrates these heterogeneous signals by computing a discounted sum of future rewards for each step, while crucially bounding these returns within the reward function's range to prevent unbounded accumulation and ensure training stability.
This enables effective credit assignment across steps with different reward objectives. That is the validation steps learn to produce accurate critiques that facilitate later corrections, while generation steps benefit from the improved validation signals.
By optimizing these complementary objectives within each sequence, the model both develops stronger self-validation capabilities and achieves better autoformalization performance, with each capability reinforcing the other throughout the training process.

Extensive experiments across four challenging autoformalization benchmarks validate the effectiveness of our \modelname. We achieve an average improvement of 22.6 percentage points over the strongest baselines.
In the choice of evaluation metrics, while recent works predominantly rely on LLMs as judges for semantic consistency evaluation in autoformalization~\citep{wang2025kimina,lin2025goedelv2}, the reliability of these LLM-based judges is not sufficiently studied.
To rigorously investigate the reliability of frontier LLMs as evaluation metrics, we construct \textbf{ConsistencyCheck}, a benchmark of 859 expert-annotated items to test how accurately a model determines whether a given formal statement correctly captures the problem's intent.
Our analysis on ConsistencyCheck reveals three insights:
(1) \textbf{Human Expert Fallibility}: 16.4\% of miniF2F and 38.5\% of ProofNet's human-written formal statements contain semantic errors, demonstrating that autoformalization challenges even human experts.
(2) \textbf{Evaluation Reliability Despite Imperfection}: Frontier LLMs make correct determination 85.8\% of the time, indicating sufficient reliability as an evaluation metric.
Crucially, \modelname's substantial improvements far exceed the potential evaluation noise, confirming the robustness of our findings.
(3) \textbf{Classification-Generation Gap}: This binary classification task is conceptually simpler than autoformalization task. However, its maximum performance is 85.8\%, which helps explain why generating semantically faithful formalizations remains stubbornly difficult.
These results confirm the effectiveness of our reflective method in producing more reliable and semantically faithful autoformalization.

In summary, our contributions are as follows:
\begin{itemize}[topsep=1pt, partopsep=1pt, leftmargin=12pt, itemsep=-1pt]

    \item We propose \modelname, a reflective autoformalization paradigm that reconceptualizes autoformalization from one-pass translation to an iterative process interweaving generation with semantic self-validation, enabling progressive error identification and correction.

    \item We introduce Prospective Bounded Sequence Optimization (PBSO) to handle different reward signals at different sequence positions through prospective bounded returns that improve both autoformalization and semantic validation.

    \item We construct ConsistencyCheck benchmark to rigorously evaluate the reliability of LLM-based metrics and the quantify the challenges autoformalization poses.

    \item We demonstrate \modelname's effectiveness across four challenging benchmarks. The model achieves an average improvement of 22.6\% while maintaining computational efficiency.

\end{itemize}

%% file: section/2.related_work.tex
\section{Related Works}

\textbf{Autoformalization.}
Autoformalization—the translation of natural language problems into machine-verifiable formal languages—plays a pivotal role in formal mathematical reasoning~\citep{wu2023autoformalization,jiang2311multilingual,ying2024lean,xie2025fmc,he2025deepmath,zhang2025deeptheorem,jiayi2025solving,yu2025formalmath}.
While early approaches achieved syntactic correctness through increasing training data~\citep{han2024lean,xin2024deepseekb}, recent studies~\citep{peng2025criticlean} reveal that these models suffer from pervasive semantic infidelity.
Recent concurrent works~\citep{wang2025kimina,lin2025goedelv2} focus on semantically-enhanced datasets to address this issue, yet they remain constrained by the \textit{one-pass generation paradigm} without correction mechanisms.
In contrast, our \modelname interweaves autoformalization with semantic self-validation, allowing the model to progressively identify and correct its own semantic errors during generation.

% In contrast, our \modelname introduces a \textit{reflective autoformalization paradigm} that interweaves generation with semantic self-validation, enabling progressive error identification and correction during the autoformalization process.

\textbf{Reinforcement Learning for LLM Reasoning.}
Reinforcement Learning (RL) has emerged as a powerful paradigm for enhancing LLM reasoning capabilities~\citep{schulman2017proximal,Shao2024DeepSeekMath,chen2024alphamath,chen2024step,Wang2025Reinforcement,Yu2025DAPO,qiao2025webresearcher,chen2025iterresearch}.
However, existing methods predominantly rely on terminal-only rewards.
While effective for single-objective tasks, this reward paradigm fails to monitor intermediate validation steps in multi-objective tasks, leading to superficial or hallucinated critiques that undermine the self-correction process.
In contrast, our PBSO introduces a \textit{prospective bounded return} to integrate heterogeneous rewards across sequence positions.
This enables position-specific optimization for both reflective autoformalization and general sequential decision-making tasks with multi-objective requirements.

%% file: section/3.method.tex
\section{Methodology}
\label{sec:method}
\begin{figure}[t]
    \centering
    \includegraphics[width=\linewidth]{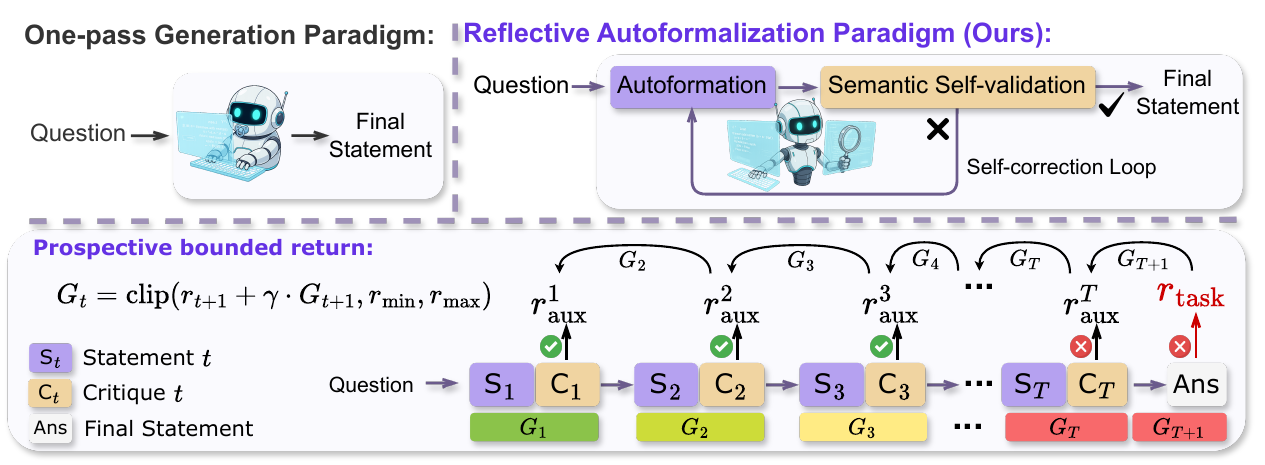}
    % \vspace{-1em}
    \caption{
    Overview of \modelname.
    \textbf{(Top)} Unlike traditional one-pass generation, our \modelname reconceptualizes it as a iterative process that interweaves autoformalization with semantic self-validation.
    \textbf{(Bottom)} We assign heterogeneous rewards across iterations: auxiliary rewards $r_{\text{aux}}^t$ for critique quality and task reward $r_{\text{task}}$ for final correctness. 
Prospective bounded returns $G_t$ computed through clipped backward accumulation enable fine-grained credit assignment for each iteration, preventing the degeneration of self-validation while improving autoformalization performance.
    }
    \label{fig:ReForm}
\end{figure}

As illustrated in Figure~\ref{fig:ReForm}, we present \modelname, our reflective autoformalization framework that departs from the prevailing one-pass generation paradigm by introducing an iterative self-correction process for enhanced semantic consistency.
In this section, we describe the Reflective Autoformalization Paradigm (\S\ref{sec:reflective}) followed by the Prospective Bounded Sequence Optimization (\S\ref{sec:pbso}).

\subsection{Reflective Autoformalization Paradigm}
\label{sec:reflective}
The core innovation of \modelname lies in reconceptualizing autoformalization as an iterative refinement process that interweaves formal statement generation with semantic self-validation.
Unlike traditional one-pass approaches, our reflective paradigm establishes a self-correction loop where the model progressively refines its output based on its own semantic critiques.

Given a natural language mathematical question $Q$, \modelname operates through a sequence of refinement iterations. 
At iteration $t$, the model maintains the complete history of previous attempts: $\mathcal{H}_t = \{(S_1, C_1), \ldots, (S_{t-1}, C_{t-1})\}$, where $S_j$ and $C_j$ denote the $j$-th formal statement and its corresponding semantic critique.
This history enables the model to learn from its previous attempts, avoiding repeated errors and progressively converging toward a semantically faithful formalization.
Each iteration comprises two interconnected stages:
\begin{enumerate}[topsep=1pt, partopsep=1pt, leftmargin=12pt, itemsep=-0.5pt]
    \item \textbf{Autoformalization:} The model $\pi$ generates a new formal statement $S_t$ conditioned on the question $Q$ and history $\mathcal{H}_t$. For the initial iteration ($t=1$), the model performs standard autoformalization: $S_1 = \pi(Q)$. For subsequent iterations ($t>1$), the generation leverages insights from previous critiques to address identified semantic issues: $S_t = \pi(Q, \mathcal{H}_t)$.

    \item \textbf{Semantic Self-Validation:} Given the newly generated statement $S_t$, the model produces a critique $C_t = \pi(Q, \mathcal{H}_t, S_t)$ that assesses the semantic consistency between $S_t$ and $Q$. The critique $C_t$ provides detailed diagnostic feedback, identifying specific semantic discrepancies, enabling targeted improvements in subsequent iterations.
\end{enumerate}
This ``Autoformalization $\leftrightarrow$ Self-validation'' loop continues until the critique indicates satisfactory semantic fidelity after which the model produces the final statement.

\textbf{Unified Generation.} %Through Autoregressive Decoding.
While conceptually iterative, our paradigm is implemented as a single continuous autoregressive generation. 
Both autoformalization and self-validation naturally unfold within the same generation sequence: the model generates $S_t$, which becomes part of the context for generating $C_t$, which in turn informs $S_{t+1}$.
This design ensures that the entire reflective process occurs within a single forward pass, maintaining computational efficiency comparable to the one-pass methods while achieving superior semantic consistency.
The model learns to autonomously identify and correct semantic errors during generation, effectively internalizing what would traditionally require multiple model calls and human oversight.

\textbf{Mutual Reinforcement with Dual Capabilities.}
The interweaving of autoformalization and self-validation creates a virtuous cycle: the autoformalization capability progressively develops semantic awareness by learning from self-validation, while the self-validation capability becomes increasingly adept at identifying subtle errors by observing refinement patterns.
This mutual reinforcement stands in contrast to traditional one-pass approaches and is instrumental in enabling \modelname to achieve formal statements that are both syntactically valid and semantically faithful.

\subsection{Prospective Bounded Sequence Optimization}
\label{sec:pbso}
While the reflective autoformalization paradigm establishes the structural framework for iterative refinement, training models to effectively execute this process presents a unique challenge: \textit{how to jointly optimize for both correct final autoformalizations and accurate intermediate self-validations within a single sequence?}

The core difficulty lies in the heterogeneous nature of these objectives. 
High-quality self-validation critiques are essential for guiding refinement, yet they serve as diagnostic tools rather than direct solutions.
A model might generate insightful critiques identifying all semantic issues but fail to translate these insights into correct refinements, or conversely, produce correct formalizations despite superficial self-validations.
This creates an important credit assignment problem: optimizing solely for final task success provides no explicit signal for the critique quality, potentially causing the self-validation mechanism to degenerate into trivial or hallucinated assessments.

\subsubsection{Heterogeneous Reward Mechanism}
To address this challenge, we introduce a heterogeneous reward structure that supervises both the primary autoformalization task and the auxiliary self-validation task:

\textbf{Task Reward for Autoformalization.}
We assign a positive reward to the final formal statement (termed `\textbf{Ans}' below) only when it achieves both syntactic and semantic correctness:
\begin{equation} \label{eq:task_reward}
    r_{\text{task}}(Q, \text{Ans}) = \begin{cases} 
        1 & \text{if } \text{\texttt{PassesLean}}(\text{Ans}) \land \text{\texttt{IsConsistent}}(Q, \text{Ans}) \\
        0  & \text{otherwise} 
    \end{cases}
\end{equation}
where \texttt{PassesLean} verifies syntactic validity through the Lean compiler, and \texttt{IsConsistent} assesses semantic consistency between ``Ans'' and the original question $Q$ using an LLM-based judge\footnote{\texttt{IsConsistent} and \texttt{IsFaithfulCritique} are evaluated by \texttt{CriticLean-14B}~\citep{peng2025criticlean} and \texttt{Qwen3-235B-A22B}~\citep{yang2025qwen3}, respectively. We provide detailed reliability evaluations in \S~\ref{sec:model_valid}.}.
This reward drives the primary learning objective, encouraging the model to produce correct final formalizations.

\textbf{Auxiliary Rewards for Self-Validation Quality.}
To prevent degeneration of self-validation, we introduce auxiliary rewards that directly supervise each critique $C_t$:
\begin{equation} \label{eq:aux_reward}
    r_{\text{aux}}^t(Q, S_t, C_t) = \begin{cases} 
        1 & \text{if } \text{\texttt{IsFaithfulCritique}}(Q, S_t, C_t) \\
        0 & \text{otherwise} 
    \end{cases}
\end{equation}
where \texttt{IsFaithfulCritique} evaluates whether $C_t$ accurately diagnoses the semantic relationship between the current statement $S_t$ and question $Q$, penalizing false positives, false negatives, and premature termination (incorrectly claiming semantic fidelity when discrepancies remain).

Together, these heterogeneous rewards provide complementary supervision signals. The task reward ensures correct final outputs while auxiliary rewards maintain the integrity of the self-validation mechanism, jointly enabling effective reflective autoformalization.

\subsubsection{Prospective Bounded Return}
Existing RL methods typically assign rewards only at sequence termination, optimizing solely for task success.
In our reflective paradigm, this method would provide little to no supervision for the quality of intermediate self-validations.
Without explicit rewards for self-validation, the model is at risk of learning to generate superficial or hallucinated critiques that appear to justify refinements but provide no genuine diagnostic value, thereby undermining the entire reflective mechanism.

To address this problem, we introduce a prospective bounded return that maximizes expected cumulative reward while ensuring quality at each step.
Our approach integrates both task and auxiliary rewards distributed across the trajectory, where each position's return $G_t$ captures the cumulative value of the remaining sequence from that point forward.
This prospective view enables the model to learn how current decisions contribute to eventual task success.
For a trajectory with $T$ iterations producing rewards $[r_{\text{aux}}^1, \ldots, r_{\text{aux}}^T, r_{\text{task}}]$, we compute returns for each step through backward accumulation with bounded discounting:
\begin{equation}
\label{eq:bounded_return}
G_t = \text{clip}\left(r_t + \gamma \cdot G_{t+1}, r_{\min}, r_{\max}\right)
\end{equation}
where $\gamma \in (0, 1]$ is the discount factor, $G_{T+1} = 0$, and the clipping operation bounds returns within the reward function's range $[r_{\min}, r_{\max}]$ to prevent gradient instability from unbounded accumulation.
Each $G_t$ serves as the composite reward signal for the entire $t$-th iteration—encompassing both the statement generation $S_t$ and its critique $C_t$—capturing how this complete reflective step contributes to the trajectory's overall success.

\subsubsection{Sequence Optimization with Position-specific Advantages}
Building on the prospective bounded returns, we now present our complete Prospective Bounded Sequence Optimization (PBSO) algorithm.
Unlike existing RL methods that compute advantages using only terminal task rewards without supervising intermediate steps, PBSO leverages the full sequence of heterogeneous returns to compute position-specific advantages.
This enables fine-grained credit assignment where each iteration receives distinct supervision based on its actual contribution to the trajectory's success.

For each question $Q$, we sample $N$ complete trajectories, where trajectory $j$ undergoes $T_j$ iterations.
The bounded return computation (Eq.~\ref{eq:bounded_return}) yields a sequence of returns $\{G_1^j, G_2^j, \ldots, G_{T_j+1}^j\}$ capturing the prospective value at each iteration.
To enable policy optimization, we transform these returns into advantages through joint normalization across all sampled trajectories:
\begin{equation}
\hat{A}_t^j = \frac{G_t^j - \text{mean}(\mathcal{G})}{\text{std}(\mathcal{G})}, \quad \text{where} \quad \mathcal{G} = \bigcup_{j=1}^N \{G_t^j : t = 1, \ldots, T_j+1\}
\end{equation}
This produces position-specific advantage sequences $\hat{\mathbf{A}}^j = [\hat{A}_1^j, \hat{A}_2^j, \ldots, \hat{A}_{T_j+1}^j]$ for each trajectory, where all tokens within iteration $t$ receive advantage $\hat{A}_t^j$.
These advantages vary across iterations even within the same trajectory—early iterations that successfully identify critical errors may receive higher advantages than later iterations that make minor refinements.
We then update the policy using these position-specific advantages with standard GRPO~\citep{Shao2024DeepSeekMath}, jointly optimizing both autoformalization accuracy and self-validation quality.

% The key innovation of PBSO is that advantages vary across iterations even within the same trajectory—early iterations that successfully identify critical errors may receive higher advantages than later iterations that make minor refinements.
% This fine-grained credit assignment enables the model to learn which specific reflective behaviors contribute most to problem-solving success.

%% file: section/4.exp.tex
\section{Experiment}

\subsection{Experimental Setup}
\label{subsec:setup}
\textbf{Datasets.}
To rigorously assess the effectiveness of \modelname, we evaluate on four challenging benchmarks:
\textbf{(1) miniF2F}~\citep{zheng2021minif2f}: 244 test problems from high-school mathematics competitions.
\textbf{(2) ProofNet}~\citep{azerbayev2023proofnet}: 186 undergraduate-level theorems from textbooks spanning real analysis, abstract algebra, and topology.
\textbf{(3) PutnamBench}~\citep{tsoukalas2024putnambench}: 644 college-level competition problems from the Putnam Mathematical Competition (1962-2023).
\textbf{(4) AIME2025}~\citep{aime25}: 30 problems from the 2025 American Invitational Mathematics Examination, testing autoformalization on contemporary competition problems.

\textbf{Baselines.}
We compare our \modelname against the state-of-the-art methods including:
\textbf{(1) Proprietary and Open-source Models}: We evaluate frontier LLMs including GPT-5~\citep{gpt_5}, Claude-3.7-Sonnet~\citep{claude_37}, Gemini-2.5-Pro~\citep{gemini_25_pro}, DeepSeek-R1-0528~\citep{Guo2025DeepSeekR1}, QwQ-32B~\citep{qwen_qwq}, and Qwen3 series~\citep{yang2025qwen3}.
\textbf{(2) Autoformalization Models}: We compare with state-of-the-art autoformalization models, including DeepSeek-Prover-V1.5-RL~\citep{xin2024deepseekb}, Goedel-V1~\citep{lin2025goedel}, Kimina-Autoformalizer-7B~\citep{wang2025kimina}, and Goedel-FormalizerV2 (8B and 32B)~\citep{lin2025goedelv2}.

\textbf{Evaluation Metrics.}
We consider two key metrics:
\textbf{(1) Syntactic Correctness (syn)}: whether the formal statement passes Lean compiler verification; 
\textbf{(2) Semantic Consistency (sem)}: whether the statement is both syntactically correct and semantically faithful. This is our primary metric.
We adopt \texttt{Qwen3-235B-A22B} as our default evaluation model unless explicitly stated otherwise.

\textbf{Implementation Details.}
We implement \modelname based on Qwen3 (8B and 32B)~\citep{yang2025qwen3}.
We curate training data from diverse open sources including Omni-MATH~\citep{gao2024omni}, IneqMath~\citep{jiayi2025solving}, the Lean Workbook~\citep{ying2024lean}, DeepTheorem~\citep{zhang2025deeptheorem}, Natural Proofs~\citep{razborov1994natural}, and Big-Math~\citep{albalak2025big}, with rigorous deduplication against all test sets to ensure fair evaluation.
For details, we refer readers to Appendix~\ref{app:imp_details} for dataset statistics, SFT and RL training procedures.

\subsection{Main Results}
\input{table/main_result}
Table~\ref{tab:main_result} presents comprehensive evaluation results across four challenging autoformalization benchmarks.
Additionally, we provide \texttt{CriticLean} based evaluation results to further validate the robustness of our findings (Appendix~\ref{app:criticlean_result}).
We have two key findings that validate the effectiveness of the reflective paradigm.

\textbf{First}, our \modelname achieves state-of-the-art performance with substantial improvements across all benchmarks.
\modelname-8B surpasses the strongest baseline Goedel-FormalizerV2-8B by an average of +14.8pp in semantic consistency while improving syntactic correctness (+8.2pp).
Remarkably, our 8B model even outperforms the 4× larger Goedel-FormalizerV2-32B by +14.2pp in semantic consistency, demonstrating that the reflective paradigm's architectural innovation goes byond mere parameter scaling.
The improvements are particularly pronounced on the more challenging benchmarks: +18.3pp on ProofNet and +14.4pp on PutnamBench compared to Goedel-V2-8B, with the most dramatic gain of +20.0pp on AIME2025. These consistent improvements across diverse benchmarks confirm that iterative self-validation fundamentally enhances semantic understanding.

\textbf{Second}, the reflective paradigm specifically addresses the critical semantic issue that plagues existing autoformalization systems.
\textbf{All baseline methods exhibit a severe syntactic-semantic performance gap}, revealing that models readily generate Lean-compilable code but systematically fail at semantic fidelity. This gap is most extreme in the more difficult datasets (e.g. Putname).
In contrast, \modelname's improvements are more concentrated on semantic consistency than on syntactic refinement.
This asymmetric improvement pattern amplifies on harder benchmarks, with semantic gains consistently exceeding syntactic improvements by 2-3×, demonstrating that by making semantic validation an integral part of generation, \modelname transforms autoformalization from superficial pattern matching to genuine mathematical understanding.
%Most tellingly, \textbf{ReForm achieves the smallest syntactic-semantic gaps while maintaining the highest absolute performance}, demonstrating that by making semantic validation an integral part of generation, our approach transforms autoformalization from superficial pattern matching to genuine mathematical understanding.

\subsection{Ablation Studies}\label{sec:ablation}
\input{table/ablation}
We conduct comprehensive ablation studies to analyze the contribution of each component in our framework, as shown in Table~\ref{tab:ablation_study}. The experiments are divided into two parts:
\textbf{(1) Training Components.}
Removing the bounded clipping in Eq.~\ref{eq:bounded_return} causes severe degradation particularly on the harder benchmarks, confirming that bounding returns is crucial for stable optimization with heterogeneous rewards.
The auxiliary reward $r_{\text{aux}}$ exhibits increasing importance as problem complexity increases, indicating that explicit supervision for self-validation quality becomes more critical for harder problems.
The RL training phase provides consistent improvements across all datasets, with gains increasing on harder problems, demonstrating that PBSO effectively learns complex reasoning strategies beyond SFT.
\textbf{(2) Paradigm Comparison.}
The most striking result emerges from comparing our reflective paradigm against one-pass generation:
when trained on identical data, the one-pass baseline shows dramatic performance gaps that widen with problem difficulty.
This widening gap validates our core hypothesis: as mathematical complexity increases, the need for iterative self-correction becomes paramount.
Single-pass generation fundamentally lacks the mechanism to identify and rectify its own semantic errors, while our reflective paradigm enables progressive refinement through self-validation.
% Notably, even without RL training, the reflective paradigm (w/o RL row) substantially outperforms the one-pass approach, confirming that the architectural innovation itself—independent of training methodology—is transformative for semantic consistency.
% The synergy between our reflective paradigm and PBSO training yields optimal performance, confirming that both architectural innovation and specialized optimization are essential for achieving faithful autoformalization.

\subsection{Training dynamics of PBSO}\label{exp:rl}
\begin{figure}[htbp]
    \centering
    \begin{subfigure}[b]{0.325\textwidth}
        \centering
        \includegraphics[width=\textwidth]{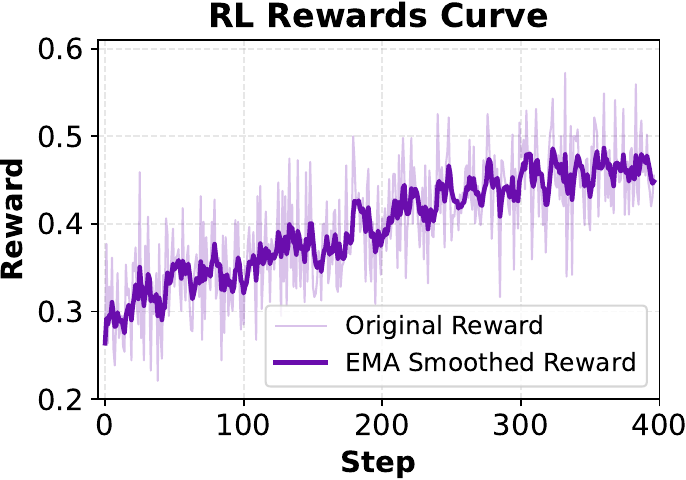}
    \end{subfigure}
    \hfill
    \begin{subfigure}[b]{0.325\textwidth}
        \centering
        \includegraphics[width=\textwidth]{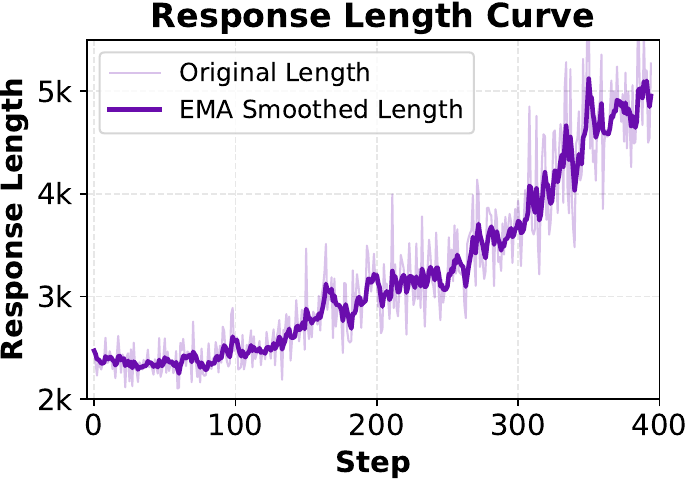}
    \end{subfigure}
    \hfill
    \begin{subfigure}[b]{0.325\textwidth}
        \centering
        \includegraphics[width=\textwidth]{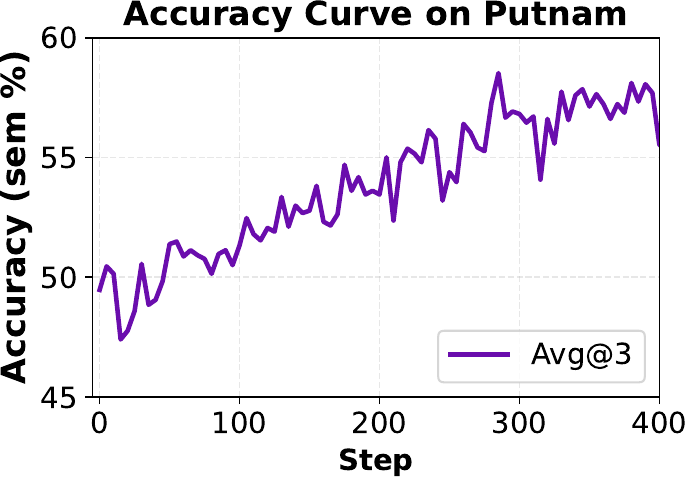}
    \end{subfigure}
    \caption{Training dynamics of our RL process.}
    \label{fig:training_dynamics}
\end{figure}
To understand how Prospective Bounded Sequence Optimization enables effective learning of the reflective paradigm, we analyze the training dynamics in \modelname-8B across three key dimensions.
Figure~\ref{fig:training_dynamics} reveals three critical insights into how PBSO shapes model behavior:

\textbf{(1) Stable Heterogeneous-Objective Optimization.} The training reward (left) steadily improves from 0.30 to 0.47 over 400 steps, with progressively narrowing confidence bands indicating stable convergence.
This smooth progression demonstrates that PBSO successfully balances heterogeneous rewards at different sequence positions, optimizing both task success and critique quality.
\textbf{(2) Emergent Reflective Behavior.} 
Response length (middle) exhibits remarkable organic growth from 2,300 to 4,800 tokens during training—a 2.1× expansion solely from heterogeneous reward signals, without any explicit length bonuses or penalties.
This phenomenon reveals a crucial insight: when properly incentivized through auxiliary rewards for critique quality, models autonomously develop more thorough self-examination behaviors.
\textbf{(3) Robust Generalization.} 
Performance on held-out PutnamBench (right) improves from 47\% to 57\% in semantic consistency, closely tracking training rewards.
This tight correlation between training and test performance, maintained throughout optimization rather than diverging due to overfitting, demonstrates that PBSO enables learning of transferable reflective capabilities.
Together, these dynamics reveal how PBSO orchestrates the RL process: heterogeneous rewards drive the emergence of reflective behavior, which in turn generates richer training signals, further improving both autoformalization and self-validation capabilities in a virtuous cycle.

\subsection{Reliability of Semantic Consistency Evaluation}\label{sec:model_valid}
Since our evaluation relies on LLM-based judges to assess semantic consistency, establishing their reliability is crucial for validating our experimental conclusions.
We construct \textbf{ConsistencyCheck}, a benchmark of 859 expert-annotated items where models perform binary classification: determining whether a formal statement correctly preserves the mathematical semantics of the original question.

\textbf{Human expert fallibility in existing benchmarks.}
During the annotation process, we uncovered that 16.4\% of miniF2F and 38.5\% of ProofNet's human-written formal statements contain semantic errors.
This high error rate in expert-crafted formalizations underscores that autoformalization challenges even human specialists, further motivating the need for automated approaches like \modelname.

\input{table/consistency}

\textbf{LLM evaluation reliability analysis.}
Table~\ref{tab:consistency} reveals that while Gemini-2.5-Pro achieves the highest accuracy (85.8\%), open-source Qwen3-235B-A22B provides comparable performance (82.9\%) with balanced precision-recall trade-offs.
These results reveal two critical insights for the autoformalization community:
(1) \textbf{Classification-Generation Gap validates autoformalization's difficulty.}
On this classification task, which is inherently simpler than generation, frontier models plateau at 86\% accuracy.
This 14\% error rate in merely \textit{recognizing} semantic consistency helps explain why \textit{generating} faithful formalizations remains fundamentally challenging, as generation requires not just recognition but creative synthesis under semantic constraints.
(2) \textbf{Current evaluation is sufficiently reliable for our experiments.}
Despite imperfections, with an accuracy at $85.8\%$, current LLMs provide adequate signals for drawing research conclusions.
Crucially, \modelname's improvements far exceed potential evaluation noise: our +14.4pp gain on PutnamBench represents a 2.5-standard-deviation effect size given the judge's error rate, while our +20.0pp improvement on AIME2025 corresponds to 3.5 standard deviations — both statistically robust.
Based on these analyses, we adopt \texttt{Qwen3-235B-A22B} as our primary semantic judge (balancing quality with reproducibility) and \texttt{CriticLean-14B} for RL training (for efficiency). Finally,
ConsistencyCheck is released to facilitate future research on autoformalization evaluation reliability.

\paragraph{Human Evaluation on \modelname}
While LLM-based evaluation provides scalable assessment, we further conduct human evaluation to directly validate \modelname's outputs.
We evaluate the final formal statements generated by \modelname-8B on miniF2F and ProofNet test sets.
Each statement was classified as ``Correct'' only if it was both syntactically valid and semantically faithful to the original problem; otherwise, it was deemed ``Incorrect''.
The human evaluation revealed high fidelity, with 86.1\% of miniF2F and 69.4\% of ProofNet formalizations verified as correct, closely aligning with our LLM-based semantic consistency scores (87.7\% and 65.6\% respectively) and validating the reliability of automated evaluation.

\subsection{Robustness Analysis on Main Result Evaluated By CriticLean-14B}\label{app:criticlean_result}
\input{table/criticlean_result}

To validate that our improvements are not artifacts of a specific evaluation metric, Table~\ref{tab:criticlean_main_result} presents results evaluated by CriticLean-14B~\citep{peng2025criticlean}, an independent semantic consistency judge trained specifically for autoformalization assessment.

\textbf{(1) Consistent Improvements Across Evaluators.} 
\modelname maintains substantial advantages under CriticLean evaluation, with ReForm-8B achieving an average improvement of +18.0pp over Goedel-V2-8B (compared to +14.8pp under Qwen3-235B evaluation).
This consistency across fundamentally different evaluators—a general-purpose LLM (Qwen3) versus a specialized critic model (CriticLean)—strongly validates the robustness of our approach.

\textbf{(2) Amplified Gains on Challenging Benchmarks.}
The improvements are even more pronounced under CriticLean evaluation for difficult datasets: +33.3pp on AIME2025 (vs +20.0pp with Qwen3) and +18.8pp on PutnamBench (vs +14.4pp).
This suggests that CriticLean may be more sensitive to semantic nuances in complex problems, where our reflective paradigm provides the greatest benefits.

\textbf{(3) Different Absolute Scores but Consistent Rankings.}
While CriticLean generally assigns higher semantic consistency scores than Qwen3 (possibly due to different training objectives or calibration), the relative rankings remain largely consistent.
Notably, \modelname achieves the highest scores under both evaluators, with ReForm-8B reaching 70.3\% average semantic consistency under CriticLean—a remarkable achievement given the inherent difficulty of autoformalization.

These results from Table~\ref{tab:main_result} and Table~\ref{tab:criticlean_main_result} confirm that the reflective paradigm's effectiveness transcends specific evaluation methodologies, providing robust improvements in semantic consistency regardless of how it is measured.

%% file: table/main_result.tex
\begin{table}[t]
\centering
\caption{Main results. We report both syntactic correctness (syn) and semantic consistency (sem), with sem being our primary metric. $^\ddagger$Improvements are relative to the best baseline with comparable model size. The best results are in \textbf{bold}, and the second best are \underline{underlined} among baselines. }
\vspace{-0.5em}
\label{tab:main_result} 
\resizebox{\linewidth}{!}{

\begin{tabular}{@{}lcccccccccc@{}}
\toprule
\multicolumn{1}{l|}{}                                 & \multicolumn{2}{c|}{\textbf{miniF2F}}                                                                    & \multicolumn{2}{c|}{\textbf{ProofNet}}                                                                    & \multicolumn{2}{c|}{\textbf{Putnam}}                                                                      & \multicolumn{2}{c|}{\textbf{AIME2025}}                                                                    & \multicolumn{2}{c}{\textbf{AVG}}                                                                         \\
\multicolumn{1}{l|}{\multirow{-2}{*}{\textbf{Model}}} & syn                                                & \multicolumn{1}{c|}{sem}                            & syn                                                & \multicolumn{1}{c|}{sem}                             & syn                                                & \multicolumn{1}{c|}{sem}                             & syn                                                & \multicolumn{1}{c|}{sem}                             & syn                                                & sem                                                 \\ \midrule
\multicolumn{11}{c}{\cellcolor[HTML]{E5E5FC}\textit{\textbf{Proprietary and Open-source Models}}}                                                                                                                                                                                                                                                                                                                                                                                                                                                                                                               \\ \midrule
\multicolumn{1}{l|}{GPT 5}                            & 70.9                                               & \multicolumn{1}{c|}{66.0}                           & 49.5                                               & \multicolumn{1}{c|}{44.6}                            & 61.6                                               & \multicolumn{1}{c|}{45.8}                            & 13.3                                               & \multicolumn{1}{c|}{13.3}                            & 48.8                                               & 42.4                                                \\
\multicolumn{1}{l|}{Claude-3.7-sonnet}                & 40.2                                               & \multicolumn{1}{c|}{34.0}                           & 28.5                                               & \multicolumn{1}{c|}{22.0}                            & 20.2                                               & \multicolumn{1}{c|}{10.1}                            & 3.3                                                & \multicolumn{1}{c|}{3.3}                             & 22.2                                               & 16.5                                               \\
\multicolumn{1}{l|}{Gemini-2.5-pro}                   & 28.7                                               & \multicolumn{1}{c|}{25.8}                           & 23.1                                                & \multicolumn{1}{c|}{8.1}                             & 15.7                                               & \multicolumn{1}{c|}{6.5}                             & 13.3                                               & \multicolumn{1}{c|}{0.0}                             & 20.2                                               & 10.1                                                 \\
\multicolumn{1}{l|}{DeepSeek-R1-0528}                 & 38.5                                               & \multicolumn{1}{c|}{35.2}                           & 11.3                                               & \multicolumn{1}{c|}{9.7}                            & 19.7                                               & \multicolumn{1}{c|}{11.3}                            & 16.7                                               & \multicolumn{1}{c|}{3.3}                             & 21.6                                               & 14.9                                                \\
\multicolumn{1}{l|}{Qwen3-235B-A22B}                  & 55.7                                               & \multicolumn{1}{c|}{43.9}                           & 16.7                                               & \multicolumn{1}{c|}{12.9}                            & 33.1                                               & \multicolumn{1}{c|}{19.9}                            & 20.0                                               & \multicolumn{1}{c|}{13.3}                            & 31.4                                               & 22.5                                                \\
\multicolumn{1}{l|}{Qwen3-32B}                        & 57.4                                               & \multicolumn{1}{c|}{53.3}                           & 10.8                                                & \multicolumn{1}{c|}{8.1}                             & 8.2                                                & \multicolumn{1}{c|}{6.2}                             & 10.0                                               & \multicolumn{1}{c|}{10.0}                            & 21.6                                               & 19.4                                                \\
\multicolumn{1}{l|}{Qwen3-8B}                         & 37.7                                               & \multicolumn{1}{c|}{31.6}                           & 7.5                                                & \multicolumn{1}{c|}{5.9}                             & 4.5                                                & \multicolumn{1}{c|}{3.1}                             & 3.3                                                & \multicolumn{1}{c|}{0.0}                             & 16.6                                               & 10.1                                                \\ \midrule
\multicolumn{11}{c}{\cellcolor[HTML]{E5E5FC}\textit{\textbf{Autoformalization Models}}}                                                                                                                                                                                                                                                                                                                                                                                                                                                                                                                         \\ \midrule
\multicolumn{1}{l|}{DeepSeek-Prover-V1.5-RL}          & 86.1                                               & \multicolumn{1}{c|}{43.0}                           & 36.6                                               & \multicolumn{1}{c|}{16.1}                            & 11.3                                               & \multicolumn{1}{c|}{8.6}                             & 0.0                                                & \multicolumn{1}{c|}{0.0}                             & 30.8                                               & 14.8                                                \\
\multicolumn{1}{l|}{Goedel-V1-32B-Workbook}       & 95.1        
           & \multicolumn{1}{c|}{47.1}                            & 48.4    
                       & \multicolumn{1}{c|}{18.3}                             & 62.1        
                                     & \multicolumn{1}{c|}{9.3}                             & 
70.0                                                 &\multicolumn{1}{c|}{3.3}                              & 68.9                                               & 19.5                                                \\
\multicolumn{1}{l|}{Goedel-V1-32B-Sonnet}            & 93.4                                              & \multicolumn{1}{c|}{69.3}                             & 47.8                                              & \multicolumn{1}{c|}{26.3}                             & 73.1
                                     & \multicolumn{1}{c|}{17.2}                            &
\underline{80.0}                                     & \multicolumn{1}{c|}{13.3}                                        & 73.6                                    & 31.5 \\
\multicolumn{1}{l|}{Kimina-Autoformalizer-7B}         & 92.6                                               & \multicolumn{1}{c|}{67.6}                           & 53.2                                               & \multicolumn{1}{c|}{23.6}                            & 69.7                                               & \multicolumn{1}{c|}{25.1}                            & \underline{80.0}                                   & \multicolumn{1}{c|}{16.7}                            & 73.9                                               & 33.3                                                \\
\multicolumn{1}{l|}{Goedel-Formalizer-V2-8B}           & \underline{97.5}                                   & \multicolumn{1}{c|}{81.1}                           & 70.4                                               & \multicolumn{1}{c|}{47.3}                            & \underline{74.5}                                   & \multicolumn{1}{c|}{\underline{42.9}}                            & 66.7                                               & \multicolumn{1}{c|}{\underline{26.7}}                            & \underline{77.3}                                               & 49.5                                                \\
\multicolumn{1}{l|}{Goedel-Formalizer-V2-32B}          & 97.1                                               & \multicolumn{1}{c|}{\underline{82.0}}               & \underline{71.5}                                   & \multicolumn{1}{c|}{\underline{50.5}}                & 74.2                                               & \multicolumn{1}{c|}{41.4}                & 66.7                                               & \multicolumn{1}{c|}{\underline{26.7}}                & \underline{77.3}                                   & \underline{50.1}                                    \\ \midrule
\multicolumn{11}{c}{\cellcolor[HTML]{E5E5FC}\textit{\textbf{Ours}}}                                                                                                                                                                                                                                                                                                                                                                                                                                                                                                                                             \\ \midrule
\multicolumn{1}{l|}{\modelname-8B}                    & \textbf{98.4}                                      & \multicolumn{1}{c|}{87.7}                           & 78.5                                               & \multicolumn{1}{c|}{65.6}                            & 81.9                                               & \multicolumn{1}{c|}{57.3}                            & 83.3                                              & \multicolumn{1}{c|}{46.7}                            & 85.5                                               & 64.3                                                \\
\multicolumn{1}{c|}{+ Improvement$^\ddagger$}                    & \multicolumn{1}{l}{$\textcolor{red}{\uparrow0.9}$} & \multicolumn{1}{l|}{$\textcolor{red}{\uparrow6.6}$} & \multicolumn{1}{l}{$\textcolor{red}{\uparrow8.1}$} & \multicolumn{1}{l|}{$\textcolor{red}{\uparrow18.3}$} & \multicolumn{1}{l}{$\textcolor{red}{\uparrow7.4}$} & \multicolumn{1}{l|}{$\textcolor{red}{\uparrow14.4}$} & \multicolumn{1}{l}{$\textcolor{red}{\uparrow3.3}$} & \multicolumn{1}{l|}{$\textcolor{red}{\uparrow20.0}$} & \multicolumn{1}{l}{$\textcolor{red}{\uparrow8.2}$} & \multicolumn{1}{l}{$\textcolor{red}{\uparrow14.8}$} \\ \midrule
\multicolumn{1}{l|}{\modelname-32B}                   & \multicolumn{1}{c}{97.1}                               & \multicolumn{1}{c|}{\textbf{91.4}}                               & \multicolumn{1}{c}{\textbf{82.3}}                               & \multicolumn{1}{c|}{\textbf{70.4}}                                & \multicolumn{1}{c}{\textbf{83.1}}                               & \multicolumn{1}{c|}{\textbf{62.3}}                                & \multicolumn{1}{c}{\textbf{86.7}}                               & \multicolumn{1}{c|}{\textbf{66.7}}                                & \multicolumn{1}{c}{\textbf{87.3}}                               & \textbf{72.7}                              \\
\multicolumn{1}{c|}{+ Improvement$^\ddagger$}                    &                          $\textcolor{red}{\uparrow0.0}$                          & \multicolumn{1}{c|}{$\textcolor{red}{\uparrow9.4}$}                               &          $\textcolor{red}{\uparrow10.8}$                                          & \multicolumn{1}{c|}{$\textcolor{red}{\uparrow19.9}$}                                &       $\textcolor{red}{\uparrow8.9}$                                             & \multicolumn{1}{c|}{$\textcolor{red}{\uparrow20.9}$}                                &       $\textcolor{red}{\uparrow6.7}$                                             & \multicolumn{1}{c|}{$\textcolor{red}{\uparrow40.0}$}                                &     $\textcolor{red}{\uparrow10.0}$                                               &       $\textcolor{red}{\uparrow22.6}$                                              \\ \bottomrule
\end{tabular}

}
% \vspace{-1.5em}
\end{table}

%% file: table/ablation.tex
\begin{wraptable}{r}{0.55\textwidth}
\centering
\footnotesize{
\vspace{-1em}
\caption{Ablation studies on training methodology and paradigm with semantic consistency score. All variants use identical training data to ensure fair comparison.}
\vspace{-1em}
\label{tab:ablation_study}
\begin{tabular}{@{}lcccc@{}}
\toprule
             Method                 & \textbf{miniF2F}    & \textbf{ProofNet}    & \textbf{Putnam}    & \textbf{AIME25}   \\ \midrule
\multicolumn{5}{c}{\cellcolor[HTML]{E5E5FC}\textit{\textbf{Ablation on training Methodology}}}                             \\ \midrule
\modelname               & \textbf{87.7}       & \textbf{65.6}        & \textbf{57.3}      & \textbf{46.7}       \\
\quad w/o clip                & 84.0                & 59.6                 & 48.9               & 26.7                \\
\quad w/o $r_{\text{aux}}$    & \textbf{87.7}       & \textbf{65.6}        & 52.1               & 40.0                \\
\quad w/o RL                & 85.2               & 62.3                 & 49.4              & 30.0                \\ \midrule
\multicolumn{5}{c}{\cellcolor[HTML]{E5E5FC}\textit{\textbf{Ablation on Paradigm}}} \\ \midrule
One-pass              & 82.7                & 59.1                 & 40.8               & 16.7                \\ \bottomrule
\end{tabular}
\vspace{-0.3em}
}
\end{wraptable}

%% file: table/consistency.tex
\begin{table}[h]
\centering
\caption{LLM performance on ConsistencyCheck benchmark for semantic consistency evaluation. $^\dagger$Full model names: Claude-3.7-Sonnet, Qwen3-235B-A22B-Thinking, CriticLean-14B.}
\vspace{-1em}
\label{tab:consistency} 
\resizebox{\linewidth}{!}{

\begin{tabular}{@{}lccccccc@{}}
\toprule
Metrics            & \textbf{GPT-5} & \textbf{Gemini-2.5-pro} & \textbf{Claude-3.7}$^\dagger$ & \textbf{DeepSeek-R1} & \textbf{Qwen3-235B}$^\dagger$ & \textbf{QwQ}     & \textbf{CriticLean}$^\dagger$ \\ \midrule
\textbf{Accuracy}  & 82.5           & \textbf{85.8}           & 77.2                & 78.1                 & \underline{82.9}    & 77.9             & 79.1                \\
\textbf{Precision} & \textbf{88.9}  & 84.4                    & 75.7                & 84.7                 & \underline{85.3}    & 75.5             & 80.7                \\
\textbf{Recall}    & 82.9           & \textbf{96.9}           & 93.3                & 79.0                 & 87.7                & \underline{95.4} & 87.3                \\
\textbf{F1}        & 85.8           & \textbf{90.2}           & 83.6                & 81.8                 & \underline{86.5}    & 84.3             & 83.9                \\ \bottomrule
\end{tabular}
% \vspace{-10em}
}
\end{table}

%% file: table/criticlean_result.tex
\begin{table}[t]
\centering
\caption{Main results. We report semantic consistency (sem) based on \texttt{CriticLean-14B}~\citep{peng2025criticlean} with sem being our primary metric. $^\ddagger$Improvements are relative to the best baseline with comparable model size. The best results are in \textbf{bold}, and the second best are \underline{underlined}.}
% \vspace{-1em}
\label{tab:criticlean_main_result} 
% \resizebox{\linewidth}{!}{

\begin{tabular}{@{}lccccc@{}}
\toprule
\textbf{Model}                               & \textbf{miniF2F}               & \textbf{ProofNet}               & \textbf{Putnam}                 & \textbf{AIME2025}               & \textbf{AVG}                    \\ \midrule
\multicolumn{6}{c}{\cellcolor[HTML]{E5E5FC}\textit{\textbf{Proprietary and Open-source Models}}}                                                                                                                      \\ \midrule
GPT-5                                        & 66.8                            & 41.5                            & 44.1                            & 13.3                            & 41.7                            \\
Claude-3.7-Sonnet                            & 34.8                           & 22.6                            & 10.9                            & 0.0                             & 17.1                            \\
Gemini-2.5-Pro                               & 28.3                           & 7.0                             & 4.5                             & 0.0                             & 10.0                            \\
DeepSeek-R1-0528                             & 33.2                           & 10.2                            & 11.2                            & 3.3                             & 14.5                            \\
Qwen3-235B-A22B                              & 44.7                           & 12.4                            & 18.8                            & 23.3                            & 24.8                            \\
Qwen3-32B                                    & 54.1                           & 7.5                             & 6.7                             & 6.7                             & 18.8                            \\
Qwen3-8B                                     & 32.0                           & 7.0                             & 3.1                             & 0.0                             & 10.5                            \\ \midrule
\multicolumn{6}{c}{\cellcolor[HTML]{E5E5FC}\textit{\textbf{Autoformalization Models}}}                                                                                                                                \\ \midrule
DeepSeek-Prover-V1.5-RL                      & 44.3                           & 0.0                             & 0.5                             & 0.0                             & 11.2                            \\
Goedel-V1-32B-Workbook                       & 48.8                           & 18.3                            & 9.6                             & 3.3                             & 20.0                            \\
Goedel-V1-32B-Sonnet                         & 66.0                           & 23.1                            & 17.4                            & 10.0                            & 29.1                            \\
Kimina-Autoformaiizer-7B                     & 66.8                           & 22.0                              & 26.2                            & 13.3                            & 32.1                            \\
Goedel-Formalizer-V2-8B                      & 86.9                           & 54.8                            & 40.8                            & 26.7                            & 52.3                            \\
Goedel-Formalizer-V2-32B                     & \underline{89.3}               & \underline{59.1}                & \underline{44.3}                & \underline{33.3}                & \underline{56.5}                \\ \midrule
\multicolumn{6}{c}{\cellcolor[HTML]{E5E5FC}\textit{\textbf{Ours}}}                                                                                                                                                    \\ \midrule
\modelname-8B                                & \textbf{92.2}                  & 69.4                            & 59.6                   & 60.0                   & 70.3                   \\
\multicolumn{1}{c}{+ Improvement$^\ddagger$} & $\textcolor{red}{\uparrow5.3}$ & $\textcolor{red}{\uparrow14.6}$ & $\textcolor{red}{\uparrow18.8}$ & $\textcolor{red}{\uparrow33.3}$ & $\textcolor{red}{\uparrow18.0}$ \\ \midrule
\modelname-32B                               & 91.4                           & \textbf{73.7}                   & \textbf{64.6}                            & \textbf{63.3}                   & \textbf{73.2}                            \\
\multicolumn{1}{c}{+ Improvement$^\ddagger$} & $\textcolor{red}{\uparrow2.1}$ & $\textcolor{red}{\uparrow14.6}$ & $\textcolor{red}{\uparrow20.3}$ & $\textcolor{red}{\uparrow30.0}$ & $\textcolor{red}{\uparrow16.7}$ \\ \bottomrule
\end{tabular}

% }
% \vspace{-1.5em}
\end{table}

%% file: section/5.conclusion.tex
\section{Conclusion}
We introduce \modelname, a reflective autoformalization paradigm that fundamentally shifts from one-pass generation to an iterative process interweaving generation with semantic self-validation.
To effectively train the reflective paradigm, we propose Prospective Bounded Sequence Optimization, which realizes heterogeneous rewards via prospective bounded returns.
This enable models to jointly optimize for both correct final formalizations and accurate intermediate critiques, preventing degenerate or hallucinated self-validations.
Extensive experiments demonstrate the effectiveness of our \modelname with an average improvement of 22.6\% across four benchmarks.
Our ConsistencyCheck benchmark further reveals that autoformalization challenges even human experts while confirming the reliability of LLM-based evaluation metrics.

%% file: section/appendix.tex
\newpage
\appendix

% \section{Statement on LLM Usage}

% In accordance with ICLR 2026's policies on Large Language Model Usage, we disclose that LLMs were used in a limited capacity during the preparation of this manuscript. Specifically, we employed LLMs solely for language polishing tasks, including grammar checking, improving sentence clarity, and refining word choices to enhance readability. At no point were LLMs used to generate research ideas, produce experimental code, analyze results, or draft entire sections of this paper. All scientific content, experimental design, theoretical contributions, and analytical insights are the original work of the authors. We take full responsibility for the accuracy and integrity of all claims, data, and conclusions presented in this work.

\section{More Implementation Details}

In this section, we provide a comprehensive implementation details of our proposed method. For additional insights and more intricate details, we refer the reader to our supplementary materials.

\subsection{Implementation Details}\label{app:imp_details}
% \paragraph{Benchmark Details.}
% (1) \textbf{miniF2F}\citep{zheng2021minif2f}: 244 test subset of Olympiad-level problems from high-school mathematics competitions including AMC, AIME, and IMO, using the corrected version from Kimina\citep{wang2025kimina}.
% (2) \textbf{ProofNet}\citep{azerbayev2023proofnet}: 371 undergraduate-level theorems with natural language proofs drawn from popular textbooks, covering diverse topics including real analysis, abstract algebra, and topology.
% (3) \textbf{PutnamBench}\citep{tsoukalas2024putnambench}: 644 college-level competition problems from the William Lowell Putnam Mathematical Competition (1962-2023), spanning algebra, analysis, number theory, geometry, combinatorics, probability, and set theory.
% (4) \textbf{AIME2025}: 30 problems from the 2025 American Invitational Mathematics Examination, representing the most recent high-school competition problems requiring sophisticated mathematical reasoning.

\input{table/params}

\paragraph{Supervised Fine-turning Phase.}
We utilize \texttt{Slime}\footnote{\url{https://github.com/THUDM/slime}} as our training framework for the initial supervised fine-tuning phase. The detailed hyper-parameters for this phase are presented in Table~\ref{tab:sft_params}.
Since current LLMs lack inherent capabilities for our iterative autoformalization paradigm, we first construct high-quality training data that demonstrates both autoformalization and semantic self-validation behaviors. We employ a multi-agent system based on \texttt{Qwen3-235B-A22B-Thinking} to generate training trajectories that embody our reflective paradigm.
For each mathematical problem in our source datasets (Section \ref{subsec:setup}), we execute the following iterative process:
\begin{itemize}[topsep=1pt, partopsep=1pt, leftmargin=12pt, itemsep=-1pt]
    \item (Step 1) \textbf{Initial Autoformalization:} Generate an initial Lean4 formalization using a standard one-pass prompt (Appendix \ref{app:prompt1}).
    \item (Step 2) \textbf{Syntactic Validation:} Verify the generated statement compiles successfully in Lean4. Non-compiling statements trigger subsequent refinement.
    \item (Step 3) \textbf{Semantic Consistency Evaluation:} For syntactically valid statements, apply a consistency checking prompt (Appendix \ref{app:prompt2}) to assess semantic alignment with the original problem.
    \item (Step 4) \textbf{Reflective Refinement:} When inconsistencies are detected, combine the failed statement with evaluation feedback to generate corrections using reflective prompts (Appendix \ref{app:prompt1}, \ref{app:prompt2}).
    \item (Step 5) \textbf{Iteration:} Repeat steps 2-4 until either (a) semantic consistency is achieved, (b) maximum iterations (3 rounds) are reached, or (c) no further improvements are generated.
\end{itemize}
This pipeline produces training trajectories that naturally interweave autoformalization attempts with self-validation and correction, providing rich supervision for learning our reflective paradigm.
The resulting dataset contains 447,508 trajectories with iteration distributions detailed in Table~\ref{tab:statistics}, where 83.1\% of problems achieve resolution within a single iteration while the remaining require multiple rounds of refinement.

\input{table/statistics}

\paragraph{Reinforcement Learning Phase.}
For training data in RL phase, we start with a diverse batch of mathematical problems spanning various difficulties, grade levels, and domains.
For each problem, we generate 8 candidate formalizations by sampling from our SFT model.
These candidates were then evaluated against two successive criteria: compiler verification and a semantic consistency check.
From this pool, we curate a final dataset with 2048 items for RL.
The selection was deliberately stratified to include problems with varying pass rates (i.e., the proportion of the eight samples that passed the checks), thereby ensuring the dataset represented a wide spectrum of formalization difficulty.
Moreover, Table~\ref{tab:rl_params} summarizes the key hyperparameters used during the reinforcement learning phase.
We also use \texttt{Slime} as our RL frameowrk due to its efficient and easy to use.

\paragraph{Inference Phase.}
During inference, we employ deterministic sampling with temperature 0.6 and top-p 0.95 to balance between generation quality and diversity. The maximum generation length is set to 40,960 tokens. We utilize vLLM\footnote{\url{https://github.com/vllm-project/vllm}} as our inference engine.

% \paragraph{Human Evaluation of \modelname}
% To validate the final outputs, we conducted a final round of human evaluation on the \texttt{formal\_statements} generated by \modelname. Each statement was classified as ``Correct'' only if it was both syntactically valid and semantically faithful to the original problem; otherwise, it was deemed ``Incorrect''. The evaluation at a pass@1 setting revealed a high degree of fidelity, with 90~\% of the \texttt{miniF2F} and 87~\% of the \texttt{ProofNet} formalizations being verified as correct. When expanding the evaluation to pass@8, we found that our human annotators could identify at least one correct formalization among the eight generated candidates for 99~\% of the \texttt{miniF2F} and 93~\% of the \texttt{ProofNet} problems.

\subsection{Impact of RL Training on Iterative Refinement Behavior}

\begin{figure}[h]
    \centering
    \includegraphics[width=0.5\linewidth]{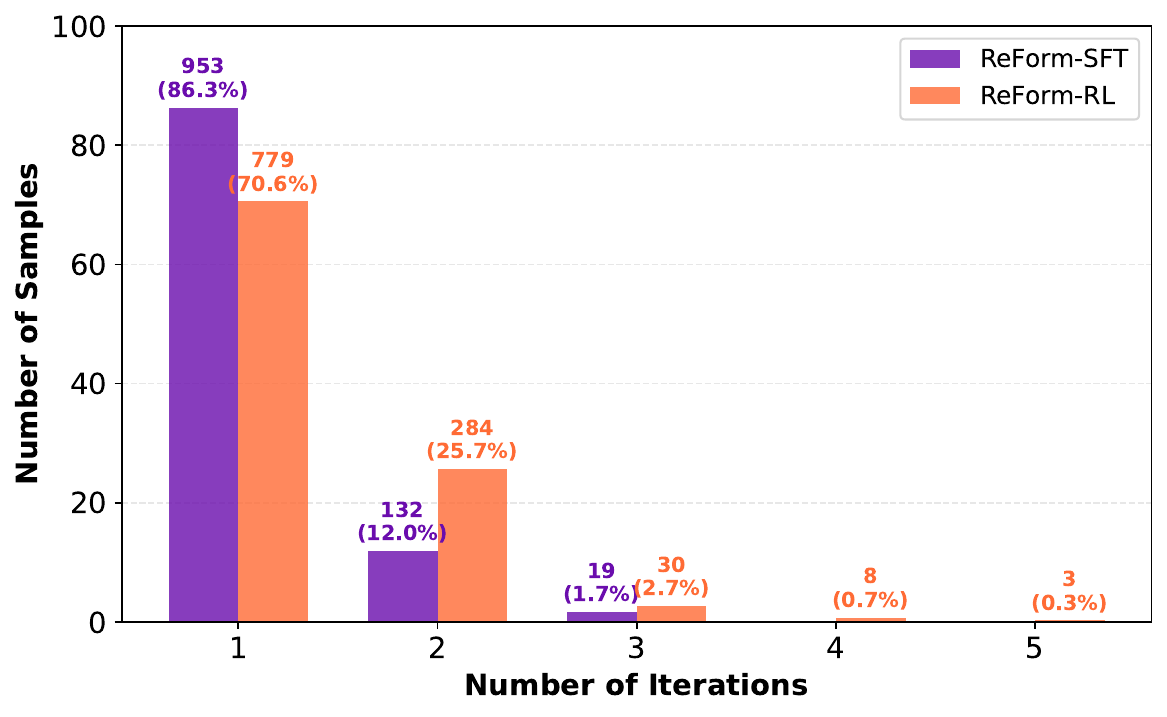}
    \caption{Iteration Distribution of our \modelname-SFT and RL.}
    \label{fig:iteration}
\end{figure}

We further analyze the distribution of iteration rounds for ReForm-8B across four benchmarks, comparing models after SFT and RL training stages. As shown in Figure~\ref{fig:iteration}, the results reveal striking behavioral changes that validate our reflective paradigm's effectiveness.

\modelname-SFT exhibits a heavily skewed distribution, with 86.3\% of samples terminating after a single iteration and the remaining samples distributed across 2-3 iterations—a pattern directly reflecting our SFT training data, which was capped at three iterations.
In contrast,\modelname-RL demonstrates a markedly different behavioral pattern. Most notably, it explores iteration depths never seen during SFT training, with 0.7\% and 0.3\% of samples extending to 4 and 5 iterations respectively. More importantly, the overall distribution shifts toward deeper refinement: the percentage of single-iteration completions drops to 70.6\%, while 2-iteration cases nearly double from 12.0\% to 25.7\%. This redistribution indicates that \textbf{PBSO training successfully teaches the model to recognize when additional refinement is beneficial, rather than prematurely terminating the reflective process}.
The emergence of 4-5 iteration trajectories—patterns entirely absent from the SFT training data—demonstrates that PBSO enables genuine exploration beyond the supervised distribution, discovering more effective refinement strategies through trial and error. This finding is further corroborated by the consistent increase in average response length during RL training (Section~\ref{exp:rl}), confirming that the model learns to invest more computational effort in challenging problems that benefit from extended reflection.

The results provide strong empirical evidence that our heterogeneous reward mechanism successfully prevents the degeneration of self-validation capabilities while encouraging productive iteration when needed, ultimately validating the core premise of our reflective autoformalization paradigm: that iterative self-correction, when properly incentivized, leads to superior semantic consistency.

\section{The ConsistencyCheck Benchmark}

\paragraph{Annotation Team.}
Our annotation team for both the ConsistencyCheck benchmark and the autoformalization results in Section 4.5 comprised 6 members. All are senior PhD. candidates with a strong background in mathematical competitions and prior experience in formalization-related annotation tasks.

\paragraph{Expertise \& Training.} To ensure high-quality and consistent annotations, a rigorous training protocol was implemented. This included dedicated sessions on interpreting the annotation guidelines, mastering the criteria for semantic consistency, and standardizing the handling of ambiguous or edge cases. Furthermore, all annotators are active researchers in Lean and formalization-related fields, possessing practical experience in formal proof development or autoformalization.

\paragraph{Annotation Protocol.} Our annotation protocol employed a three-annotator design per statement to ensure robustness. Initially, two annotators worked independently. Subsequently, a third senior annotator reviewed their annotations and accompanying textual comments (examples of which are provided in Appendix C) to perform cross-validation, resolve any discrepancies, and render the final judgment.

\paragraph{Annotation process.}
To construct a high-quality benchmark for evaluating semantic consistency, we commission a team of experts with deep proficiency in both mathematics and Lean4. The annotators were tasked with assessing the semantic fidelity of formal statements from a dataset composed of items from miniF2F~\citep{zheng2022miniff} and ProofNet~\citep{azerbayev2023proofnet}, which were presented to them in an anonymized format.
For each item, \textbf{two experts} independently compared the machine-verifiable formal statement against the original natural language problem. If the formalization faithfully captured the problem's semantic intent, it was labeled as ``Correct''. If any semantic discrepancy was found, it was labeled as ``Incorrect'', and the annotators were required to provide a detailed written justification. In cases of disagreement, \textbf{a third senior expert} was brought in to adjudicate and determine the final label, ensuring the reliability of our benchmark.

\paragraph{Example of our ConsistencyCheck Benchmark.}
A representative example of a semantic error discovered in a ProofNet entry is illustrated in Example~\ref{box:consistency_examples}. In this instance, the human-authored formalization contained two critical flaws: (1) a ``constant term mismatch'', where $\sqrt{11}$ from the natural language was incorrectly transcribed as $11$; and (2) a ``degree bound inconsistency'', where the constraint ``degree $\le 80$'' was altered to ``degree $< 80$''. 
Notably, the annotation team also identified another significant class of error in miniF2F entries: cases where the formal statement included an explicit answer, while the original natural language problem did not. These instances were also systematically classified as ``Incorrect.'' Such discrepancies underscore the profound difficulty of achieving true semantic fidelity, even in expert-curated datasets.

Based on this benchmark, we investigate the performance of each LLMs in semantic consistency evaluation. These models are tasked with assessing the semantic consistency following the prompt specified in Appendix \ref{app:prompt2}, as discussed in Section~\ref{sec:model_valid}.

% To gauge their performance, we compare the models' automated assessments against our expert human evaluations as shown in Section~\ref{sec:model_valid}, which serve as the ground truth. 

\input{prompts/case1}

\section{Case Study of our \modelname}

We present a detailed case study demonstrating our model's reflective autoformalization capabilities in practice. The example below illustrates how \modelname successfully identifies and corrects its own semantic errors through iterative self-validation.

\textbf{Round 1: Initial Attempt and Error Detection.} The model generates its first autoformalization attempt, then performs semantic self-validation by systematically analyzing the formalization across five critical dimensions: (1) variable definitions and types, (2) mathematical constraints and conditions, (3) logical relationships between statements, (4) completeness of the formalization, and (5) alignment with the problem's intent. Through this structured analysis, the model autonomously identifies semantic inconsistencies in its initial output (highlighted in \textcolor{red}{red}).

\textbf{Round 2: Self-Correction and Validation.} Leveraging its self-generated critique as feedback, the model produces a refined autoformalization that directly addresses the identified issues. The corrected statement undergoes another round of semantic validation, this time passing all consistency checks. This successful self-correction demonstrates the model's ability to not only detect its own errors but also synthesize this understanding into improved formalizations.

This example showcases the core strength of our reflective paradigm: the model's capacity for genuine self-improvement without external intervention, achieving semantic correctness through iterative refinement guided by its own critical analysis.

% As illustrated in the example below, this process unfolds in distinct rounds. In Round 1, the model first generates an initial autoformalization. It then proceeds to Semantic Self-validation of round 1, where it critically analyzes its own output across five distinct aspects and successfully identifies its own errors (highlighted in red). In Round 2, the model leverages this self-generated critique as direct feedback to produce a revised autoformalization. This refined statement successfully corrects the initial mistakes, ultimately yielding a semantically correct formal statement that passes the subsequent round 2 semantic self-validation check, thus completing a full cycle of self-reflection.

% Training on our structured dataset, which explicitly models the iterative refinement process, enables our model to execute a coherent reflective workflow at inference time. As illustrated in the example below, this process unfolds in distinct rounds. In Round 1, the model first generates an initial autoformalization. It then proceeds to Semantic Self-validation of round 1, where it critically analyzes its own output across five distinct aspects and successfully identifies its own errors (highlighted in red). In Round 2, the model leverages this self-generated critique as direct feedback to produce a revised autoformalization. This refined statement successfully corrects the initial mistakes, ultimately yielding a semantically correct formal statement that passes the subsequent round 2 semantic self-validation check, thus completing a full cycle of self-reflection.

\input{prompts/case2}
\section{Prompt}

\subsection{Instruction for Collecting SFT Trajectories}

The autoformalization prompt firstly initializes the reflective data construction process by converting a natural language mathematical problem into a Lean 4 formal statement. 

When an initial formalization fails the semantic consistency check, this prompt takes as input: (1) the original mathematical problem, (2) the failed formal statement, and (3) the detailed feedback from the consistency evaluation explaining the semantic discrepancies. By explicitly conditioning on both the failure case and its diagnosis, the prompt guides the model to generate targeted corrections that directly address the identified issues rather than producing entirely new attempts that might introduce different errors.

This prompt operationalizes our core insight that models can learn from their own mistakes through structured self-reflection.

\input{prompts/reflective_autoformalization}\label{app:prompt1}
 
\subsection{Instruction for Semantic Consistency Check}
This section presents the semantic consistency evaluation prompt that serves as the foundation for both our training and evaluation procedures. Specifically, this prompt template:
\begin{itemize}[topsep=1pt, partopsep=1pt, leftmargin=12pt, itemsep=-1pt]
    \item Powers the \texttt{IsConsistent} reward function during RL training, providing binary semantic correctness signals
    \item Drives the consistency evaluation in our ConsistencyCheck benchmark, ensuring uniform assessment criteria
    \item A crucial component of SFT Trajectories by validating the outputs from both the initial and reflective autoformalization attempts.
\end{itemize}

\input{prompts/reflective_consistency_check}
\label{app:prompt2}

\subsection{Instruction for IsFaithfulCritique in RL}

\input{prompts/IsFaithfulCritique}
\label{app:prompt3}

\subsection{Instruction for our \modelname}

After collecting multi-turn autoformalization trajectories through our multi-agent system (Appendix~\ref{app:imp_details}), we restructure these trajectories into a unified format for SFT. The key insight is to present the entire reflective process—including initial attempts, validation results, and iterative refinements—as a single model response. This allows us to train models to internalize the complete reflective paradigm within their generation process.

\input{prompts/reform}
\label{app:prompt4}

%% file: table/params.tex
\begin{table}[h]
\centering
\begin{minipage}{0.48\textwidth}
    \centering
    \caption{Key hyperparameters in the SFT phase.}
    \label{tab:sft_params}
    \begin{tabular}{@{}lc@{}}
    \toprule
    \textbf{Hyperparameter} & \textbf{Value} \\ \midrule
    Learning Rate           & 1e-5           \\
    Batch size              & 512            \\
    \#Epochs                & 3              \\
    Chat template           & \texttt{Qwen}  \\
    Max Context Length      & 40960          \\
    Warmup ratio            & 0.03           \\
    LR scheduler type       & Cosine         \\ \bottomrule
    \end{tabular}
\end{minipage}
\hfill
\begin{minipage}{0.48\textwidth}
    \centering
    \caption{Key hyperparameters in the RL phase.}
    \label{tab:rl_params}
    \begin{tabular}{@{}lc@{}}
    \toprule
    \textbf{Hyperparameter}         & \textbf{Value} \\ \midrule
    Learning Rate                   & 1e-6           \\
    Batch size                      & 32             \\
    Group size per Question ($G$)   & 16             \\
    Temperature                     & 1.0            \\
    Top-p                          & 0.95           \\
    KL coefficient ($\lambda$)      & 0.0            \\
    Entropy coefficient             & 0.0            \\
    % Max Context Length              & 40960          \\
    \bottomrule
    \end{tabular}
\end{minipage}
\end{table}

%% file: table/statistics.tex
\begin{table}[h]
\centering
\caption{Data Statistics for SFT and RL Phase.} % 标题更精确
\vspace{-0.5em}
\label{tab:statistics} 

\begin{tabular}{@{}cccccc@{}}
\toprule
\multirow{2}{*}{\textbf{Stage}} & \multirow{2}{*}{\textbf{Data Type}} & \multicolumn{3}{c}{\textbf{Iteration Distribution}} & \multirow{2}{*}{\textbf{Total}} \\
                                &                                     & Iteration 1     & Iteration 2     & Iteration 3     &                                 \\ \midrule
SFT                             & Trajectories                        & 371,679         & 65,734          & 10,095          & 447,508                         \\
RL                              & Question only                       & -               & -               & -               & 2,048                           \\ \bottomrule
\end{tabular}
\end{table}

%% file: prompts/case1.tex
\begin{tcolorbox}[breakable,title=Examples in ConsistencyCheck Benchmark]\label{box:consistency_examples}
\colorbox{gray!70}{%
    \parbox{\dimexpr\textwidth-2\fboxsep}{%
    \centering
        \textbf{Example 1}%
    }%
}

\textbf{set:} ProofNet \\
\textbf{name:} exercise\_5\_4\_3 \\
\textbf{split:} test \\
\textbf{question:} If $a \in \mathbb{C}$ is such that $p(a) = 0$, where $p(x) = x^5 + \sqrt{2}x^3 + \sqrt{5}x^2 + \sqrt{7}x + \sqrt{11}$, show that $a$ is algebraic over $\mathbb{Q}$ of degree at most 80. \\
\textbf{header:}
\begin{lstlisting}[style=lean]
import Mathlib
open Fintype Set Real Ideal Polynomial
open scoped BigOperators
\end{lstlisting} 

\textbf{formal statement:} 
\begin{lstlisting}[style=lean, mathescape=true]
theorem exercise\_5\_4\_3 {a : $\mathbb{R}$} {p : $\mathbb{R} \rightarrow \mathbb{R}$} \n  (hp : p = $\lambda$ x, x\^{}5 + real.sqrt 2 * x\^{}3 + real.sqrt 5 * x\^{}2 + \n  real.sqrt 7 * x + 11)\n  (ha : p a = 0) : \n  $\exists$ p : polynomial $\mathbb{R}$, p.degree < 80 $\land$ a $\in$ p.roots $\land$ \n  $\forall$ n : p.support, $\exists$ a b : $\mathbb{Z}$, p.coeff n = a / b :=
\end{lstlisting}

\textcolor{red}{\textbf{Human check}}\\
Incorrect

\textcolor{red}{\textbf{Human comments}}\\
1. Polynomial constant term mismatch: $\sqrt{11}$ in natural language vs 11 in formalization. \\
2. Degree bound inconsistency: degree $\leq 80$ in natural language vs $< 80$ (i.e. $\leq 79$) in formalization. \\
These discrepancies make the formalization mathematically inequivalent to the original statement. \\
\\
\colorbox{gray!70}{%
    \parbox{\dimexpr\textwidth-2\fboxsep}{%
    \centering
        \textbf{Example 2}%
    }%
}
\\
\textbf{set:} miniF2F \\
\textbf{name:} aime\_1991\_p6 \\
\textbf{split:} test \\
\textbf{question:} Suppose $r$ is a real number for which
\[
\left\lfloor r + \frac{19}{100} \right\rfloor + \left\lfloor r + \frac{20}{100} \right\rfloor + \left\lfloor r + \frac{21}{100} \right\rfloor + \cdots + \left\lfloor r + \frac{91}{100} \right\rfloor = 546.
\]
Find $\lfloor 100r \rfloor$. (For real $x$, $\lfloor x \rfloor$ is the greatest integer less than or equal to $x$.) Show that it is 743.
 \\
\textbf{header:} 
\begin{lstlisting}[style=lean]
import Mathlib\nimport Aesop\n\nset_option maxHeartbeats 0\n\nopen BigOperators Real Nat Topology Rat\n\n
\end{lstlisting} 

\textbf{formal statement:}
\begin{lstlisting}[style=lean, mathescape=true]
theorem aime_1991_p6 (r : $\mathbb{R}$) (h$_{0}$ : ($\sum$ k in Finset.Icc (19 : $\mathbb{N}$) 91, Int.floor (r + k / 100)) = 546) :
  Int.floor (100 * r) = 743 := by
\end{lstlisting}

\textcolor{red}{\textbf{Human check}}\\
Correct
% \textcolor{red}{\textbf{Human comments}}\\
\end{tcolorbox}

%% file: prompts/case2.tex
\begin{tcolorbox}[breakable,title=Example of Our \modelname]
\textcolor{black}{\textbf{Question:}}
What value of $x$ will give the minimum value of $x^2 - 14x + 3$? Show that it is 7.\\

\textcolor{red}{\textbf{The response of our \modelname:}}
\\
\textbf{$<$think$>$} \\
\colorbox{gray!70}{%
    \parbox{\dimexpr\textwidth-2\fboxsep}{%
    \centering
        \textbf{Round 1 (Autoformalization)}%
    }%
}
\textbf{$<$round$>$} \\
The problem is asking for the value of $x$ that minimizes the expression $x^2 - 14x + 3$, and it states that this value is 7.

To formalize this in Lean 4, I need to express that:
1. The function $f(x) = x^2 - 14x + 3$ has a minimum at $x = 7$
2. This means that for all real numbers $x$, we have $f(7) \leq f(x)$

Let me think about how to express this properly in Lean 4:
\\
\begin{lstlisting}[style=lean, mathescape=true]
```lean4 
theorem test_problem : $\forall$ x : $\mathbb{R}$, (7 : $\mathbb{R}$)^2 - 14 * 7 + 3 $\leq$ x^2 - 14 * x + 3 := by sorry 
``` 
\end{lstlisting}
Wait, let me think about this more carefully. The problem is asking "what value of x will give the minimum value". This suggests we want to find the argmin of the function. 

Actually, let me be more precise. The statement should be that 7 is the point where the function achieves its minimum. So we want:
\\
\begin{lstlisting}[style=lean, mathescape=true]
``` lean4
theorem test_problem : $\forall$ x : $\mathbb{R}$, (7 : $\mathbb{R}$)^2 - 14 * 7 + 3 $\leq$ x^2 - 14 * x + 3 := by  sorry
```
\end{lstlisting}

But this is a bit awkward with the explicit casting. Let me try:
\\
\begin{lstlisting}[style=lean, mathescape=true]
``` lean4
theorem test_problem : ∀ x : ℝ, 7^2 - 14 * 7 + 3 ≤ x^2 - 14 * x + 3 := by  sorry
```
\end{lstlisting}

Actually, in Lean 4, when we write
\begin{lstlisting}[style=lean, mathescape=true]
`7^2` where `7 : ℝ`
\end{lstlisting}
it should automatically be interpreted as a real number. But to be safe, let me be explicit:
\\
\begin{lstlisting}[style=lean]
``` lean4
theorem test_problem : ∀ x : ℝ, (7 : ℝ)^2 - 14 * (7 : ℝ) + 3 ≤ x^2 - 14 * x + 3 := by  sorry
```
\end{lstlisting}

I'll verify the calculation to ensure accuracy. The key is to confirm that 7 truly minimizes the quadratic expression by checking the algebraic properties of the function. \\

Expanding the squared term reveals the minimum occurs precisely at 7, with a value of -46. The algebraic manipulation shows $x^2 - 14x + 3$ reaches its lowest point when $x = 7$, supporting the original claim about the minimum value. \\

The Lean theorem captures this relationship, demonstrating that for all real numbers, the function's value at 7 is less than or equal to the function's value at any other point.Looking at this problem, I need to formalize the statement that x = 7 gives the minimum value for the expression x² - 14x + 3.
\\

\begin{lstlisting}[style=lean]
``` lean4
import Mathlib
import Aesop

set\_option maxHeartbeats 0

open BigOperators Real Nat Topology Rat

theorem test_problem : ∀ x : ℝ, (7 : ℝ)^2 - 14 * 7 + 3 ≤ x^2 - 14 * x + 3 := by  sorry  
```
\end{lstlisting}

\colorbox{gray!70}{%
    \parbox{\dimexpr\textwidth-2\fboxsep}{%
    \centering
        \textbf{Round 1 (Semantic Self-validation)}%
    }%
}

It's time to assess if the Formal Statement correctly captures the essence of the given math problem. \\

1. Math Assertion Analysis: The problem asks for the value of $x$ that minimizes the quadratic function $x^2 - 14x + 3$ and requires proving that this value is 7. The key components are: (a) the function $f(x) = x^2 - 14x + 3$, (b) the variable $x \in \mathbb{R}$, (c) the claim that $x = 7$ is the minimizer, and (d) the requirement to prove this claim.\\

2. Lean Statement Analysis: The theorem asserts $\forall x : \mathbb{R}$, $(7^2 - 14 \cdot 7 + 3) \leq x^2 - 14x + 3$. This captures the inequality that the function's value at $x = 7$ is less than or equal to its value at any other $x \in \mathbb{R}$. However, it does not explicitly state that $x = 7$ is the minimizer (i.e., the argument where the minimum occurs). \textcolor{red}{The theorem only asserts the inequality, not the existence of the minimizer or its uniqueness.}\\

3. Comparative Verification: The Lean statement correctly captures the inequality part of the problem (that f(7) is the minimum value). However, it does not explicitly state that x=7 is the value that achieves this minimum. The problem requires both identifying the minimizer (x=7) and proving the inequality. The Lean theorem omits the explicit identification of x=7 as the minimizer, focusing only on the inequality.\\

4. Conclusion: \textcolor{red}{The Lean statement is incomplete as a formalization of the problem}. While it correctly captures the inequality that f(7) is the minimum value, it does not explicitly state that x=7 is the value that achieves this minimum. The problem requires both the identification of the minimizer and the proof of the inequality, but the theorem only addresses the latter.\\

5. Accuracy Confirmation: The Lean statement is incorrect. \textcolor{red}{It fails to explicitly state that x=7 is the minimizer, which is a critical part of the problem.} The theorem only asserts the inequality, not the identification of the minimizer.\\

Therefore, the outcome of the consistency verification is:\\
Incorrect\\

\textbf{$<$/round$>$}  \\

\colorbox{gray!70}{%
    \parbox{\dimexpr\textwidth-2\fboxsep}{%
    \centering
        \textbf{Round 2 (Autoformalization)}%
    }%
}

Time to reformulate the Lean 4 Formal Statement based on verification analysis. \\

\textbf{$<$round$>$}  \\

\#\# Thought Process \\

Looking at the informal statement and the critic, \textcolor{red}{I see the issue with the previous formalization}:\\

1. The problem asks for two things:
   - What value of x minimizes x² - 14x + 3
   - Show that this value is 7\\

2. The previous formalization only captured the second part (that 7 gives the minimum value) but didn't explicitly state that 7 is the value that minimizes the expression.\\

3. To properly formalize this, I need to:
   - Define the function f(x) = x² - 14x + 3
   - State that 7 is the unique minimizer of this function\\
   
4. In Lean, a proper formalization would need to express both that 7 is a minimizer and that it's the unique minimizer.\\

Given New Formal Statement:\\

\begin{lstlisting}[style=lean]
``` lean4
import Mathlib.Analysis.Calculus.Deriv.Basic
import Mathlib.Data.Real.Basic

theorem test_problem : 
  let f : ℝ → ℝ := λ x => x^2 - 14*x + 3;
  (∀ x : ℝ, f 7 ≤ f x) ∧ (∀ x : ℝ, f x = f 7 → x = 7) := by  sorry
```
\end{lstlisting}

\colorbox{gray!70}{%
    \parbox{\dimexpr\textwidth-2\fboxsep}{%
    \centering
        \textbf{Round 2 (Semantic Self-validation)}%
    }%
}
Now I'll check if the Formal Statement completely and correctly formalizes the mathematical question. \\

1. Math Assertion Analysis: The problem asks for the value of $x$ that minimizes the quadratic function $f(x) = x^2 - 14x + 3$ and requires proving that this value is 7. The mathematical structure involves: (a) defining the function $f: \mathbb{R} \to \mathbb{R}$, (b) asserting that $x = 7$ is the global minimum (i.e., $\forall x, f(7) \leq f(x)$), and (c) ensuring uniqueness (i.e., if $f(x) = f(7)$, then $x = 7$). The problem implicitly relies on the function's convexity (since the coefficient of $x^2$ is positive), guaranteeing a unique minimum.\\

2. Lean Statement Analysis: The theorem defines $f: \mathbb{R} \to \mathbb{R}$ as $x^2 - 14x + 3$. The conclusion asserts two claims: (a) $\forall x, f(7) \leq f(x)$ (global minimum), and (b) $\forall x, f(x) = f(7) \to x = 7$ (uniqueness). The function definition, domain ($\mathbb{R}$), and logical structure match the problem's requirements.\\

3. Comparative Verification: \textcolor{red}{The Lean statement correctly captures both the existence and uniqueness of the minimum at $x = 7$. The function definition matches the mathematical text.} The universal quantifiers ($\forall x$) and implications ($\to$) align with the problem's logical structure. The use of $\mathbb{R}$ ensures the domain is preserved. No extraneous conditions or missing components are present.\\

4. Conclusion: \textcolor{red}{The Lean statement accurately formalizes the problem's requirements.}\\

5. Accuracy Confirmation: The Lean statement correctly formalizes the problem. The function definition, domain, and logical claims (global minimum and uniqueness) align perfectly with the mathematical text.\\

Consequently, the outcome of the consistency check is:\\

Correct\\

\textbf{$<$/round$>$} \\

\textbf{$<$/think$>$} \\

\colorbox{gray!70}{%
    \parbox{\dimexpr\textwidth-2\fboxsep}{%
    \centering
        \textbf{Final Statement}%
    }%
}

With successful verification, here's the final Lean 4 Formal Statement:\\
\begin{lstlisting}[style=lean]
``` lean4
import Mathlib.Analysis.Calculus.Deriv.Basic
import Mathlib.Data.Real.Basic

theorem test_problem : 
  let f : ℝ → ℝ := λ x => x^2 - 14*x + 3;
  (∀ x : ℝ, f 7 ≤ f x) ∧ (∀ x : ℝ, f x = f 7 → x = 7) := by  sorry
``` 
\end{lstlisting}

\end{tcolorbox}

%% file: prompts/reflective_autoformalization.tex
\begin{tcolorbox}[title=Reflective Autoformalization for Collecting SFT trajectories, width=\linewidth, breakable]
% \begin{small}
% \textcolor{red}{Instruction for Reflectove Autoformalization}\\
You are an expert mathematician and Lean 4 programmer. 
Your task is to translate the given mathematical problem in natural language into formal statement in Lean4 syntax, strictly following the guidelines below. \\

\textcolor{red}{Guidelines} \\
   - If the previous iteration of autoformation and consistency check results are provided, analysis it and revise the autoformation according to it. \\
   - If the math problem is algebra question, the answer will be provided as well. \\
   - For non-standard mathematical problems, first reformulate them into standard mathematical format. \\
   - For complex problems containing multiple independent parts, break them down into separate theorems. \\
   - Must strictly follow Lean 4 syntax and utilize standard Lean 4 mathematical library Mathlib4 components when possible. \\
   1. Use correct type declarations and notation conventions \\
   2. Include necessary imports \\
   3. Leverage existing mathlib4 definitions and theorems \\
   4. Follow proper naming conventions \\
   - Pay attention to the consistency between the Natural Language Statement and the Formal Statement in Lean4: \\
   1. Variable domains (e.g., $\mathbb{N}$, $\mathbb{Z}$, $\mathbb{R}$, $\mathbb{R}_+$) \\
   2. Boundary conditions (especially for special values like 0,1) \\
   3. Quantifier scopes ($\forall$, $\exists$) \\
   4. Prerequisites and assumptions \\
   5. Logical implications ($\rightarrow$, $\leftrightarrow$, $\wedge$, $\vee$) \\
   6. Function types and properties \\
   7. Set-theoretic notations \\
   - Only generate the translation. Do not try to solve or prove the problem. \\
   - Include clear documentation comments for theorems. \\

\textcolor{red}{Input Format} \\
   Informal Statement: [Natural language description of math problem] \\
   History of Formal Statement and consistency Comments: [The existing Lean 4 formalization and its critique, or an empty string if none exists.] \\

\textcolor{red}{Thought Process} \\
   - Formatted Mathematical Problem \\
     1. List all known conditions \\
     2. Define variables and their domains \\
     3. State assumption \\
   - For multiple sub-problems: \\
     Sub-problem 1: [Description] \\
     Sub-problem 2: [Description] \\
     ... \\
   - For revised autofomalization \\
   1. Analyze ambiguities with natural language and potential mismatches \\
   2. Evaluate consistency comments' validity \\
   3. Identify missing assumptions or incorrect type signatures \\
   4. Determine required mathlib imports in Lean 4 \\
   5. Preserve original theorem name unless invalid \\

\textcolor{red}{Output Format} \\
   Given New Formal Statement should always use \verb|```|lean4 to start the code block and \verb|```| to end it: \\
\verb|```|lean4 \\
{[Corrected Lean4 code]} \\
\verb|```| \\

Now! It's your turn to generate the Formal Statement. \\
Informal Statement: \textcolor{blue}{\{INFORMAL STATEMENT (QUESTION) HERE.\}} \\
History Formal Statement and critics: \textcolor{blue}{\{HISTORY HERE.\}} \\
Thought: \\

\textcolor{blue}{The model's output here.}

% \end{small}
\end{tcolorbox}

%% file: prompts/reflective_consistency_check.tex
\begin{tcolorbox}[title=Instruction for Consistency Check,width=\linewidth, breakable]
% \begin{small}
% \textcolor{red}{Instruction for Reflective Consistency Check}\\
Your role is a Lean4 expert, please help me check consistency between natural language expression and its Lean4 formal statement. \\

\textcolor{red}{Guidelines for Consistency Check}  \\
1. Core Checking Requirements:  \\
- When a critique from a previous autoformalization and consistency check result is provided, you must first analyze its findings and then assess their problems. \\
- Must carefully compare the Natural Language Statement and the Formal Statement in Lean4 through a rigorous and explicit process. \\
- Determine if the Lean theorem statement is an exact and faithful formalization of the mathematical problem \\
- If any result is Incorrect of consistency, briefly list all inconsistencies and reasons leading to the Incorrect determination in comments \\

\textcolor{red}{Evaluation Stages} \\
1. Math Assertion Analysis \\
Identify all structurally and semantically relevant components of the mathematical problem, including variables, types, quantifiers, constraints, logic structure, conclusion, and so on. The analysis should be based on the actual content of the text. \\

2. Lean Statement Analysis \\
Extract all structurally and semantically relevant components from the Lean statement, including \\
- Variable domains (e.g., real numbers vs positive real numbers) \\
- Boundary conditions (especially for 0,1) \\
- Quantifier scopes \\
- Prerequisites and assumptions \\
- Logical implications \\

3. Comparative Verification \\
Check for exact correspondence between the math and Lean statements; you may refer to aspects like: \\
- Semantic alignment, logic structure, and quantifier correctness. \\
- Preservation of constraints and boundary assumptions. \\
- Accurate typing and use of variables. \\
- Syntactic validity and proper Lean usage (free from errors). \\
- Use of symbols and constructs without semantic drift. \\
- No missing elements, no unjustified additions, and no automatic corrections or completions. \\

4. Final Judgement \\
Based solely on the above analysis, judge whether the Lean statement is a correct and exact formalization of the mathematical problem. \\
- When a critique from a previous consistency check is provided, you must first analyze its findings and then assess their correctness.
- Result must be strictly "Correct" or "Incorrect" \\
- Use "Correct" ONLY when 100\% mathematical equivalence is confirmed \\

5. Accuracy Confirmation \\
If correct: clearly confirm why all elements match. \\
If incorrect: list all mismatches and explain how each one affects correctness. \\

\textcolor{red}{Intput Format} \\
The Natural Language Statement: \\
{[A math problem in Natural language]} \\
The Formal Statement in Lean4: \\
\verb|```|lean4 \\
{[A Lean 4 theorem statement formalizing the problem]} \\
\verb|```|\\
Previous round of autoformalization and sematic validation if provided: \\
{[The existing critique, or an empty string if none exists.]}
\\

\textcolor{red}{Output Format} \\
Return exactly one XML object \\
$<$comments$>$ \\
Your brief analysis: \\
1. Math Assertion Analysis: {[...]} \\
2. Lean Statement Analysis (Proof Ignored): {[...]} \\
3. Comparative Verification: {[...]} \\
4. Conclusion: {[...]} \\
5. Accuracy Confirmation: {[...match confirmation or list of discrepancies...]} \\
$<$/comments$>$ \\
$<$consistency$>$ Correct/Incorrect$<$/consistency$>$ \\
\\
Now! It's your turn to compare the natural language statement with the formal statement in Lean4: \\
The Natural Language Statement\: \textcolor{blue}{\{INFORMAL STATEMENT (QUESTION) HERE.\}} \\

The Formal Statement in Lean4: \\
\verb|```|lean4  \\
\textcolor{blue}{\{FORMAL STATEMENT HERE.\}} \\
\verb|```| \\

Previous autoformalization and consistency Comments: \\
\textcolor{blue}{\{HISTORY CRITIC HERE\}} \\
\\
Think about the consistent result:\\

\textcolor{blue}{The model's output here.}
% \end{small}
\end{tcolorbox}

%% file: prompts/IsFaithfulCritique.tex
\begin{tcolorbox}[title=Instruction for IsFaithfulCritique,width=\linewidth, breakable]
% \begin{small}
% \textcolor{red}{Instruction for Consistency Check}\\
Your role is a Lean4 expert, helping me review the previous consistency checking results. 
\\

\textcolor{red}{Guidelines for IsFaithfulCritique Check} \\
- Thoroughly examine the previous consistency result (Correct/Incorrect) and comments.
- Carefully evaluate whether the comparison results between informal statement and formal statement in previous consistency comments are correct, and if the reasoning is sufficient.
- Try to identify any errors in previous consistency comments.
- Try to compare the mathematical problem and the Lean4 formal statement to see if you would reach the same consistency conclusion.
- If you agree with the previous consistency result, provide your consistency result as "Correct"; if you disagree, provide "Incorrect" and explain your reasons for disagreement in the comments.

\textcolor{red}{Intput Format}\\
The Natural Language Statement: \\
{[A math problem in Natural language]} \\
Previous Validation result: \\
{[Previous Lean4 formal statement and semantic validation result]}\\

\textcolor{red}{Output Format}\\
Return exactly one xml object \\
\textless comments\textgreater  \\
Brief analysis of my semantic validation result, with improvements if needed. If previous result is not faithfull, explain why. \\
\textless /comments\textgreater \\
\textless consistency\textgreater Correct/Incorrect\textless /consistency\textgreater \\

Now review my consistency checking result: \\
The Natural Language Statement: \\
\textcolor{blue}{\{INFORMAL STATEMENT (QUESTION) HERE.\}}  \\

History formal statement and consistency check result:\\
\textcolor{blue}{\{PREVIOUS CONSISTENCY CHECKING COMMENTS HERE\}}

\textcolor{blue}{\{The model’s output here.\}}
% \end{small}
\end{tcolorbox}

%% file: prompts/reform.tex
\begin{tcolorbox}[breakable,title=Instruction for our Reform]

Think step by step to translate the mathematical problem in natural language to Lean 4, and verify the consistency. \\
\textcolor{blue}{\{informal\_statement\}}\\

\textcolor{blue}{The model’s output here.}
\end{tcolorbox}

%% file: iclr2026_conference.bib
@inproceedings{de2015lean,
  title={The Lean theorem prover (system description)},
  author={De Moura, Leonardo and Kong, Soonho and Avigad, Jeremy and Van Doorn, Floris and von Raumer, Jakob},
  booktitle={International Conference on Automated Deduction},
  pages={378--388},
  year={2015},
  organization={Springer}
}

@article{wu2023autoformalization,
  title={Autoformalization with large language models},
  author={Wu, Yilun and Jiang, Albert Q and Li, Wenda and Jamnik, Mateja and Lample, Guillaume and Rabe, M},
  journal={Transactions on Machine Learning Research},
  year={2023}
}

@article{jiang2311multilingual,
  title={Multilingual mathematical autoformalization},
  author={Jiang, Albert Qiaochu and Li, Wenda and Jamnik, Mateja},
  journal={arXiv preprint arXiv:2311.03755},
  year={2023}
}

@article{han2024lean,
  title={Lean workbook: A large-scale lean problem set formalized from natural language math problems},
  author={Han, Chuanyang and Huang, Pu and Wang, Yixuan and Chen, Yutao and Zhang, Sheng and Song, Li},
  journal={arXiv preprint arXiv:2404.14813},
  year={2024}
}

@inproceedings{polu2020generative,
  title={Generative language modeling for automated theorem proving},
  author={Polu, Stanislas and Sutskever, Ilya},
  booktitle={International Conference on Machine Learning},
  pages={7872--7882},
  year={2020},
  organization={PMLR}
}

@article{yang2023leandojo,
  title={Leandojo: Theorem proving with retrieval-augmented language models},
  author={Yang, Kaiyu and Swope, Aidan and Gu, Alex and Chalamala, Rahul and Song, Peiyang and Yu, Shixing and Godil, Saad and Prenger, Ryan J and Anandkumar, Animashree},
  journal={Advances in Neural Information Processing Systems},
  volume={36},
  pages={21573--21612},
  year={2023}
}

@misc{team2023internlm,
  title={Internlm: A multilingual language model with progressively enhanced capabilities},
  author={{InternLM Team}},
  year={2023},
  howpublished = {arXiv preprint arXiv:2307.16135}
}

@article{wu2024internlm2,
  title={Internlm2. 5-stepprover: Advancing automated theorem proving via expert iteration on large-scale lean problems},
  author={Wu, Zijian and Huang, Suozhi and Zhou, Zhejian and Ying, Huaiyuan and Wang, Jiayu and Lin, Dahua and Chen, Kai},
  journal={arXiv preprint arXiv:2410.15700},
  year={2024}
}

@inproceedings{wu2025internlm2,
  title={InternLM2. 5-StepProver: Advancing Automated Theorem Proving via Critic-Guided Search},
  author={Wu, Zijian and Huang, Suozhi and Zhou, Zhejian and Ying, Huaiyuan and Yuan, Zheng and Zhang, Wenwei and Lin, Dahua and Chen, Kai},
  booktitle={2nd AI for Math Workshop@ ICML 2025},
  year={2025}
}

@article{xin2024deepseek,
  title={Deepseek-prover: Advancing theorem proving in llms through large-scale synthetic data},
  author={Xin, Huajian and Guo, Daya and Shao, Zhihong and Ren, Zhizhou and Zhu, Qihao and Liu, Bo and Ruan, Chong and Li, Wenda and Liang, Xiaodan},
  journal={arXiv preprint arXiv:2405.14333},
  year={2024}
}

@article{xin2024deepseekb,
  title={Deepseek-prover-v1. 5: Harnessing proof assistant feedback for reinforcement learning and monte-carlo tree search},
  author={Xin, Huajian and Ren, ZZ and Song, Junxiao and Shao, Zhihong and Zhao, Wanjia and Wang, Haocheng and Liu, Bo and Zhang, Liyue and Lu, Xuan and Du, Qiushi and others},
  journal={arXiv preprint arXiv:2408.08152},
  year={2024}
}

@article{ren2025deepseek,
  title={Deepseek-prover-v2: Advancing formal mathematical reasoning via reinforcement learning for subgoal decomposition},
  author={Ren, ZZ and Shao, Zhihong and Song, Junxiao and Xin, Huajian and Wang, Haocheng and Zhao, Wanjia and Zhang, Liyue and Fu, Zhe and Zhu, Qihao and Yang, Dejian and others},
  journal={arXiv preprint arXiv:2504.21801},
  year={2025}
}

@article{lin2025goedel,
  title={Goedel-prover: A frontier model for open-source automated theorem proving},
  author={Lin, Yong and Tang, Shange and Lyu, Bohan and Wu, Jiayun and Lin, Hongzhou and Yang, Kaiyu and Li, Jia and Xia, Mengzhou and Chen, Danqi and Arora, Sanjeev and others},
  journal={arXiv preprint arXiv:2502.07640},
  year={2025}
}

@article{wang2025kimina,
  title={Kimina-prover preview: Towards large formal reasoning models with reinforcement learning},
  author={Wang, Haiming and Unsal, Mert and Lin, Xiaohan and Baksys, Mantas and Liu, Junqi and Santos, Marco Dos and Sung, Flood and Vinyes, Marina and Ying, Zhenzhe and Zhu, Zekai and others},
  journal={arXiv preprint arXiv:2504.11354},
  year={2025}
}

@article{chen2025seed,
  title={Seed-Prover: Deep and Broad Reasoning for Automated Theorem Proving},
  author={Chen, Luoxin and Gu, Jinming and Huang, Liankai and Huang, Wenhao and Jiang, Zhicheng and Jie, Allan and Jin, Xiaoran and Jin, Xing and Li, Chenggang and Ma, Kaijing and others},
  journal={arXiv preprint arXiv:2507.23726},
  year={2025}
}

@article{qwen_qwq,
  title={Qw{Q}: Reflect deeply on the boundaries of the unknown},
  author={Qwen},
  journal={Hugging Face},
  year={2024}
}

@article{zheng2021minif2f,
  title={Minif2f: a cross-system benchmark for formal olympiad-level mathematics},
  author={Zheng, Kunhao and Han, Jesse Michael and Polu, Stanislas},
  journal={arXiv preprint arXiv:2109.00110},
  year={2021}
}

@article{tsoukalas2024putnambench,
  title={Putnambench: Evaluating neural theorem-provers on the putnam mathematical competition},
  author={Tsoukalas, George and Lee, Jasper and Jennings, John and Xin, Jimmy and Ding, Michelle and Jennings, Michael and Thakur, Amitayush and Chaudhuri, Swarat},
  journal={Advances in Neural Information Processing Systems},
  volume={37},
  pages={11545--11569},
  year={2024}
}

@article{yu2025formalmath,
  title={Formalmath: Benchmarking formal mathematical reasoning of large language models},
  author={Yu, Zhouliang and Peng, Ruotian and Ding, Keyi and Li, Yizhe and Peng, Zhongyuan and Liu, Minghao and Zhang, Yifan and Yuan, Zheng and Xin, Huajian and Huang, Wenhao and others},
  journal={arXiv preprint arXiv:2505.02735},
  year={2025}
}

@article{peng2025criticlean,
  title={CriticLean: Critic-Guided Reinforcement Learning for Mathematical Formalization},
  author={Peng, Zhongyuan and Yao, Yifan and Ma, Kaijing and Guo, Shuyue and Li, Yizhe and Zhang, Yichi and Zhang, Chenchen and Zhang, Yifan and Yu, Zhouliang and Li, Luming and others},
  journal={arXiv preprint arXiv:2507.06181},
  year={2025}
}

@article{lin2025goedelv2,
  title={Goedel-Prover-V2: Scaling Formal Theorem Proving with Scaffolded Data Synthesis and Self-Correction},
  author={Lin, Yong and Tang, Shange and Lyu, Bohan and Yang, Ziran and Chung, Jui-Hui and Zhao, Haoyu and Jiang, Lai and Geng, Yihan and Ge, Jiawei and Sun, Jingruo and others},
  journal={arXiv preprint arXiv:2508.03613},
  year={2025}
}

@article{Wang2025Reinforcement,
    author = {Wang, Weixun and Xiong, Shaopan and Chen, Gengru and Gao, Wei and Guo, Sheng and He, Yancheng and Huang, Ju and Liu, Jiaheng and Li, Zhendong and Li, Xiaoyang and others},
    title = {Reinforcement learning optimization for large-scale learning: An efficient and user-friendly scaling library},
    year = {2025}
}

@article{schulman2017proximal,
  title={Proximal policy optimization algorithms},
  author={Schulman, John and Wolski, Filip and Dhariwal, Prafulla and Radford, Alec and Klimov, Oleg},
  journal={arXiv preprint arXiv:1707.06347},
  year={2017}
}

@article{Yu2025DAPO,
    author = {Yu, Qiying and Zhang, Zheng and Zhu, Ruofei and Yuan, Yufeng and Zuo, Xiaochen and Yue, Yu and Fan, Tiantian and Liu, Gaohong and Liu, Lingjun and Liu, Xin and others},
    title = {DAPO: An open-source LLM reinforcement learning system at scale},
    year = {2025},
    eprint = {2503.14476},
    archivePrefix = {arXiv},
    primaryClass = {cs.LG}
}

@article{Guo2025DeepSeekR1,
    author = {Guo, Daya and Yang, Dejian and Zhang, Haowei and Song, Junxiao and Zhang, Ruoyu and Xu, Runxin and Zhu, Qihao and Ma, Shirong and Wang, Peiyi and Bi, Xiao and others},
    title = {DeepSeek-R1: Incentivizing reasoning capability in LLMs via reinforcement learning},
    year = {2025},
    eprint = {2501.12948},
    archivePrefix = {arXiv},
    primaryClass = {cs.CL}
}

@article{Shao2024DeepSeekMath,
    author = {Shao, Zhihong and Wang, Peiyi and Zhu, Qihao and Xu, Runxin and Song, Junxiao and Bi, Xiao and Zhang, Haowei and Zhang, Mingchuan and Li, Y. K. and Wu, Y. and others},
    title = {DeepSeekMath: Pushing the limits of mathematical reasoning in open language models},
    year = {2024},
    eprint = {2402.03300},
    archivePrefix = {arXiv},
    primaryClass = {cs.AI}
}

@article{yang2025qwen3,
  title={Qwen3 technical report},
  author={Yang, An and Li, Anfeng and Yang, Baosong and Zhang, Beichen and Hui, Binyuan and Zheng, Bo and Yu, Bowen and Gao, Chang and Huang, Chengen and Lv, Chenxu and others},
  journal={arXiv preprint arXiv:2505.09388},
  year={2025}
}

@article{gpt_4o,
  author       = {Aaron Hurst and
                  Adam Lerer and
                  Adam P. Goucher and
                  Adam Perelman and
                  Aditya Ramesh and
                  Aidan Clark and
                  AJ Ostrow and
                  Akila Welihinda and
                  Alan Hayes and
                  Alec Radford and
                  Aleksander Madry and
                  Alex Baker{-}Whitcomb and
                  Alex Beutel and
                  Alex Borzunov and
                  Alex Carney and
                  Alex Chow and
                  Alex Kirillov and
                  Alex Nichol and
                  Alex Paino and
                  Alex Renzin and
                  Alex Tachard Passos and
                  Alexander Kirillov and
                  Alexi Christakis and
                  Alexis Conneau and
                  Ali Kamali and
                  Allan Jabri and
                  Allison Moyer and
                  Allison Tam and
                  Amadou Crookes and
                  Amin Tootoonchian and
                  Ananya Kumar and
                  Andrea Vallone and
                  Andrej Karpathy and
                  Andrew Braunstein and
                  Andrew Cann and
                  Andrew Codispoti and
                  Andrew Galu and
                  Andrew Kondrich and
                  Andrew Tulloch and
                  Andrey Mishchenko and
                  Angela Baek and
                  Angela Jiang and
                  Antoine Pelisse and
                  Antonia Woodford and
                  Anuj Gosalia and
                  Arka Dhar and
                  Ashley Pantuliano and
                  Avi Nayak and
                  Avital Oliver and
                  Barret Zoph and
                  Behrooz Ghorbani and
                  Ben Leimberger and
                  Ben Rossen and
                  Ben Sokolowsky and
                  Ben Wang and
                  Benjamin Zweig and
                  Beth Hoover and
                  Blake Samic and
                  Bob McGrew and
                  Bobby Spero and
                  Bogo Giertler and
                  Bowen Cheng and
                  Brad Lightcap and
                  Brandon Walkin and
                  Brendan Quinn and
                  Brian Guarraci and
                  Brian Hsu and
                  Bright Kellogg and
                  Brydon Eastman and
                  Camillo Lugaresi and
                  Carroll L. Wainwright and
                  Cary Bassin and
                  Cary Hudson and
                  Casey Chu and
                  Chad Nelson and
                  Chak Li and
                  Chan Jun Shern and
                  Channing Conger and
                  Charlotte Barette and
                  Chelsea Voss and
                  Chen Ding and
                  Cheng Lu and
                  Chong Zhang and
                  Chris Beaumont and
                  Chris Hallacy and
                  Chris Koch and
                  Christian Gibson and
                  Christina Kim and
                  Christine Choi and
                  Christine McLeavey and
                  Christopher Hesse and
                  Claudia Fischer and
                  Clemens Winter and
                  Coley Czarnecki and
                  Colin Jarvis and
                  Colin Wei and
                  Constantin Koumouzelis and
                  Dane Sherburn},
  title        = {GPT-4o System Card},
  journal      = {CoRR},
  volume       = {abs/2410.21276},
  year         = {2024},
  url          = {https://doi.org/10.48550/arXiv.2410.21276},
  doi          = {10.48550/ARXIV.2410.21276},
  eprinttype    = {arXiv},
  eprint       = {2410.21276},
  bibsource    = {dblp computer science bibliography, https://dblp.org}
}

@miscf{claude_37,
    title={Claude 3.7 Sonnet and Claude Code},
    author={Anthropic},
    howpublished={\url{https://www.anthropic.com/news/claude-3-7-sonnet}},
    month={February},
    year={2025}
}

@miscf{gpt_5,
    title={GPT-5 is here},
    author={OpenAI},
    howpublished={\url{https://openai.com/gpt-5/}},
    month={August},
    year={2025}
}

@miscf{gemini_25_pro,
    title={Gemini 2.5 Pro},
    author={Google},
    howpublished={\url{https://deepmind.google/technologies/gemini/pro/}},
    month={April},
    year={2025}
}

@inproceedings{
zheng2022miniff,
title={miniF2F: a cross-system benchmark for formal Olympiad-level mathematics},
author={Kunhao Zheng and Jesse Michael Han and Stanislas Polu},
booktitle={International Conference on Learning Representations},
year={2022},
url={https://openreview.net/forum?id=9ZPegFuFTFv}
}

@article{azerbayev2023proofnet,
  title={Proofnet: Autoformalizing and formally proving undergraduate-level mathematics},
  author={Azerbayev, Zhangir and Piotrowski, Bartosz and Schoelkopf, Hailey and Ayers, Edward W and Radev, Dragomir and Avigad, Jeremy},
  journal={arXiv preprint arXiv:2302.12433},
  year={2023}
}

@article{albalak2025big,
  title={Big-math: A large-scale, high-quality math dataset for reinforcement learning in language models},
  author={Albalak, Alon and Phung, Duy and Lile, Nathan and Rafailov, Rafael and Gandhi, Kanishk and Castricato, Louis and Singh, Anikait and Blagden, Chase and Xiang, Violet and Mahan, Dakota and others},
  journal={arXiv preprint arXiv:2502.17387},
  year={2025}
}

@article{gao2024omni,
  title={Omni-math: A universal olympiad level mathematic benchmark for large language models},
  author={Gao, Bofei and Song, Feifan and Yang, Zhe and Cai, Zefan and Miao, Yibo and Dong, Qingxiu and Li, Lei and Ma, Chenghao and Chen, Liang and Xu, Runxin and others},
  journal={arXiv preprint arXiv:2410.07985},
  year={2024}
}

@article{jiayi2025solving,
    author = {Jiayi, Sheng and Luna, Lyu and Jikai, Jin and Tony, Xia and Alex, Gu and James, Zou and Pan, Lu},
    title = {Solving Inequality Proofs with Large Language Models},
    journal = {arXiv preprint arXiv:2506.07927},
    year = {2025}
}

@article{qiao2025webresearcher,
  title={Webresearcher: Unleashing unbounded reasoning capability in long-horizon agents},
  author={Qiao, Zile and Chen, Guoxin and Chen, Xuanzhong and Yu, Donglei and Yin, Wenbiao and Wang, Xinyu and Zhang, Zhen and Li, Baixuan and Yin, Huifeng and Li, Kuan and others},
  journal={arXiv preprint arXiv:2509.13309},
  year={2025}
}

@article{ying2024lean,
  title={Lean workbook: A large-scale lean problem set formalized from natural language math problems},
  author={Ying, Huaiyuan and Wu, Zijian and Geng, Yihan and Wang, Jiayu and Lin, Dahua and Chen, Kai},
  journal={Advances in Neural Information Processing Systems},
  volume={37},
  pages={105848--105863},
  year={2024}
}

@article{he2025deepmath,
  title={Deepmath-103k: A large-scale, challenging, decontaminated, and verifiable mathematical dataset for advancing reasoning},
  author={He, Zhiwei and Liang, Tian and Xu, Jiahao and Liu, Qiuzhi and Chen, Xingyu and Wang, Yue and Song, Linfeng and Yu, Dian and Liang, Zhenwen and Wang, Wenxuan and others},
  journal={arXiv preprint arXiv:2504.11456},
  year={2025}
}

@article{xie2025fmc,
  title={FMC: Formalization of Natural Language Mathematical Competition Problems},
  author={Xie, Jiaxuan and Liu, Chengwu and Yuan, Ye and Li, Siqi and Xiao, Zhiping and Zhang, Ming},
  journal={arXiv preprint arXiv:2507.11275},
  year={2025}
}

@article{chen2025iterresearch,
  title={Iterresearch: Rethinking long-horizon agents via markovian state reconstruction},
  author={Chen, Guoxin and Qiao, Zile and Chen, Xuanzhong and Yu, Donglei and Xu, Haotian and Zhao, Wayne Xin and Song, Ruihua and Yin, Wenbiao and Yin, Huifeng and Zhang, Liwen and others},
  journal={arXiv preprint arXiv:2511.07327},
  year={2025}
}

@article{zhang2025deeptheorem,
  title={Deeptheorem: Advancing llm reasoning for theorem proving through natural language and reinforcement learning},
  author={Zhang, Ziyin and Xu, Jiahao and He, Zhiwei and Liang, Tian and Liu, Qiuzhi and Li, Yansi and Song, Linfeng and Liang, Zhenwen and Zhang, Zhuosheng and Wang, Rui and others},
  journal={arXiv preprint arXiv:2505.23754},
  year={2025}
}

@inproceedings{razborov1994natural,
  title={Natural proofs},
  author={Razborov, Alexander A and Rudich, Steven},
  booktitle={Proceedings of the twenty-sixth annual ACM symposium on Theory of computing},
  pages={204--213},
  year={1994}
}

@misc{aime25,
  author       = {{OpenCompass}},
  title        = {{AIME 2025 Benchmark}},
  year         = {2025},
  howpublished = {\url{https://huggingface.co/datasets/opencompass/AIME2025}},
  note         = {},
}

@inproceedings{
    chen2024alphamath,
    title={AlphaMath Almost Zero: Process Supervision without Process},
    author={Guoxin Chen and Minpeng Liao and Chengxi Li and Kai Fan},
    booktitle={The Thirty-eighth Annual Conference on Neural Information Processing Systems},
    year={2024},
    url={https://openreview.net/forum?id=VaXnxQ3UKo}
}

@article{chen2024step,
  title={Step-level value preference optimization for mathematical reasoning},
  author={Chen, Guoxin and Liao, Minpeng and Li, Chengxi and Fan, Kai},
  journal={arXiv preprint arXiv:2406.10858},
  year={2024}
}

@article{li2024survey,
  title={A survey on deep learning for theorem proving},
  author={Li, Zhaoyu and Sun, Jialiang and Murphy, Logan and Su, Qidong and Li, Zenan and Zhang, Xian and Yang, Kaiyu and Si, Xujie},
  journal={arXiv preprint arXiv:2404.09939},
  year={2024}
}

@article{weng2025autoformalization,
  title={Autoformalization in the Era of Large Language Models: A Survey},
  author={Weng, Ke and Du, Lun and Li, Sirui and Lu, Wangyue and Sun, Haozhe and Liu, Hengyu and Zhang, Tiancheng},
  journal={arXiv preprint arXiv:2505.23486},
  year={2025}
}
